\documentclass{article}

\PassOptionsToPackage{numbers, compress}{natbib}

\usepackage[preprint]{neurips_2025}

\usepackage[utf8]{inputenc}   
\usepackage[T1]{fontenc}      
\usepackage{xcolor}           
\usepackage{microtype}        
\usepackage{graphicx}         
\usepackage{url}              
\usepackage{hyperref}         

\usepackage{amsmath}
\usepackage{amssymb}
\usepackage{amsthm}           
\usepackage{amsfonts}
\usepackage{mathtools}        
\usepackage{nicefrac}         
\usepackage{dsfont}           
\usepackage{booktabs}         
\usepackage{enumitem}         
\usepackage{wrapfig}

\usepackage{subcaption}       

\usepackage{algorithm}
\usepackage{algorithmic}

\theoremstyle{plain}
\newtheorem{theorem}{Theorem}[section]

\theoremstyle{definition}
\newtheorem{definition}[theorem]{Definition}

\theoremstyle{remark}

\newcommand{\cS}{\mathcal{S}}
\newcommand{\cA}{\mathcal{A}}
\newcommand{\mdp}{M}

\newcommand{\trainsitionSet}{\mathfrak{P}}
\newcommand{\expect}{\mathbb{E}}
\newcommand{\dataset}{\mathcal{D}}
\newcommand{\transition}{P_\mathcal{T}}

\title{HWC-Loco: A Hierarchical Whole-Body Control Approach to Robust Humanoid Locomotion}

\author{%
  Sixu Lin$^{1,2}$ \quad
  Guanren Qiao$^{1}$ \quad
  Yunxin Tai$^{3}$ \quad
  Ang Li$^{1,4}$ \quad
  Kui Jia$^{1,3}$ \quad
  Guiliang Liu$^{1}$\thanks{$^\dagger$Corresponding author. Email: \texttt{liuguiliang@cuhk.edu.cn}} \\
  $^1$The Chinese University of Hong Kong (Shenzhen) $\quad$
  $^2$Harbin Institute of Technology \\
  $^3$DexForce Technology $\quad$ $^4$Southeast University
}

\begin{document}

\maketitle

\begin{abstract}
Humanoid robots, capable of assuming human roles in various workplaces, have become essential to embodied intelligence. However, as robots with complex physical structures, learning a control model that can operate robustly across diverse environments remains inherently challenging, particularly under the discrepancies between training and deployment environments. In this study, we propose HWC-Loco, a robust whole-body control algorithm tailored for humanoid locomotion tasks. By reformulating policy learning as a robust optimization problem, HWC-Loco explicitly learns to recover from safety-critical scenarios. While prioritizing safety guarantees, overly conservative behavior can compromise the robot's ability to complete the given tasks. To tackle this challenge, HWC-Loco leverages a hierarchical policy for robust control. This policy can dynamically resolve the trade-off between goal-tracking and safety recovery, guided by human behavior norms and dynamic constraints. To evaluate the performance of HWC-Loco, we conduct extensive comparisons against state-of-the-art humanoid control models, demonstrating HWC-Loco's superior performance across diverse terrains, robot structures, and locomotion tasks under both simulated and real-world environments. Our project page is available at \href{https://simonlinsx.github.io/HWC_Loco/}{\texttt{HWC-Loco}}.
\end{abstract}

\vspace{-0.15in}
\section{Introduction}
\vspace{-0.05in}
\begin{wrapfigure}{r}{0.5\textwidth}
  \centering
  \vspace{-25pt}  
  \includegraphics[width=0.5\textwidth]{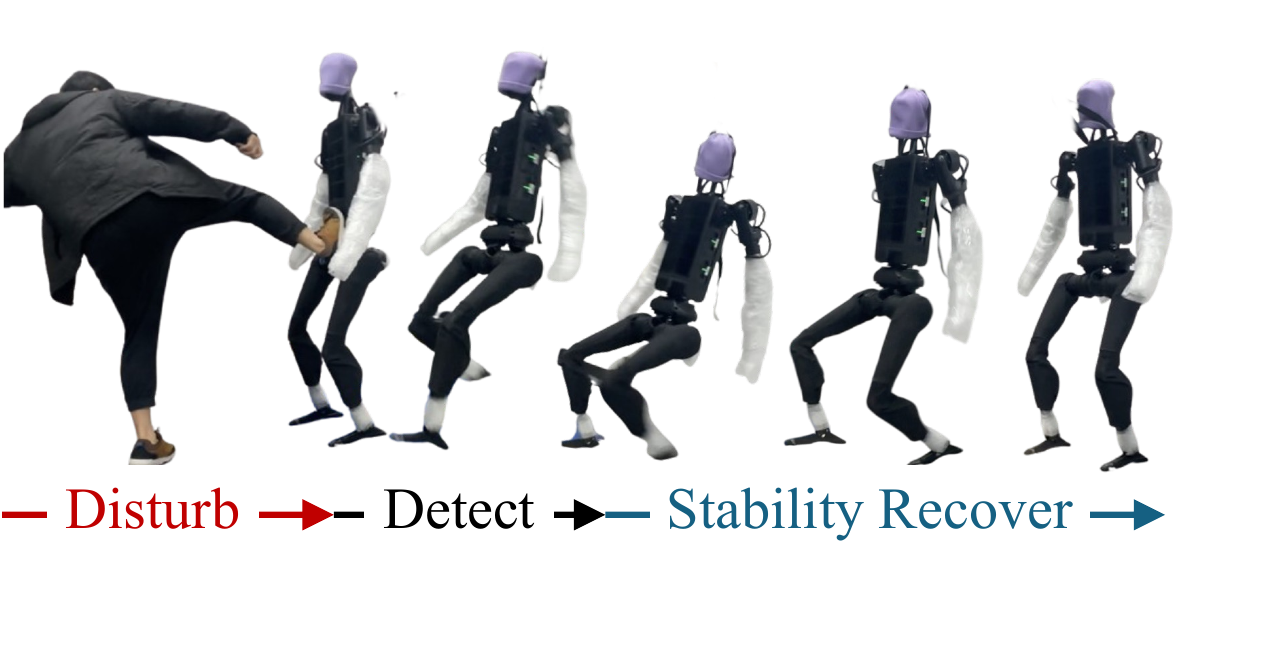}
  \vspace{-30pt} 
  \caption{An example of recovering from a Hard Kick: The humanoid robot withstands external disturbance by automatically detecting hazardous states and adjusting its motion to regain stability.}\label{fig:kick}
  \vspace{-15pt}  
\end{wrapfigure}
Humanoid robots, with a physical structure resembling that of humans, can be seamlessly integrated into humans' workspace and take on roles in completing different tasks~\cite{Saeedvand2019survey}. 
This capability makes them reliable embodiments of artificial intelligence in various forms~\cite{Paolo2024EmbodiedAI}. In recent years, with advancements in both hardware capabilities and control algorithms, humanoid robots have become an increasingly significant type of robot with a growing impact on practical applications across various environments, such as factories, homes, and offices~\cite{sferrazza2024humanoidbench,Nasiriany2024RoboCasa,tao2024maniskill3}. 


In the development of humanoid control, classic methods commonly rely on model-based optimization \cite{Sakagami2002ASIMO,Sentis2006Humanoids,Gouaillier2009humanoid,Radford2015Humanoid,chignoli2021humanoid}. However, these methods require precise and comprehensive modeling of the robot's structure, kinematics, and dynamics across different environments. In practice, obtaining such data requires significant computational resources or manual effort. These limitations significantly influence the scalability of these approaches. To develop an end-to-end solution with promising generalizability, recent studies~\cite{Zhuang2024Parkour,radosavovic2024real,he2024PIM,he2024h2o,Gu2024Locomotion} adopted Reinforcement Learning (RL) methods by training a neural model to control based on human demonstrations and interactions with the environment.

While these learning-based approaches show potential for effective control policies across various tasks, they are typically trained in simulated environments, which often differ considerably from the deploying environment. To bridge the Simulation-to-Real (Sim2Real) gap, previous works often introduce additional regularization to constrain the robot's movements or employ domain randomization to account for variations in the robot's physical properties~\cite{lai2023sim,rudin2022learning,nahrendra2023dreamwaq}
. However, excessive regularization can greatly affect the efficiency of control policy, and unstructured randomization often fails to capture safety-critical patterns in real-world applications. 

To develop a reliable locomotion policy capable of generalizing from the training to the deployment environment, we propose formulating policy optimization as a robust optimization problem under misspecified environmental dynamics.
However, traditional robust optimization primarily focuses on worst-case control, requiring the policy to ensure safety under the most adverse environmental dynamics within an uncertainty set. Under this context, the learned policy tends to be overly conservative, inducing sub-optimal control performance in terms of finishing the given task. 

In this paper, to learn a control policy that can dynamically solve the trade-off between task performance and maintaining safety under different deployment environments,  we present \textbf{H}ierarchical \textbf{W}hole-body \textbf{C}ontrol for Robust
Humanoid \textbf{Loco}motion (HWC-Loco).
Figure~\ref{fig:kick} illustrates an example of HWC-Loco, where a {\it high-level planning policy} detects hazardous states and seamlessly switches from the {\it goal-tracking policy} to a {\it safety-recovery policy}.

The {\it goal-tracking policy} is designed to follow task-specific commands (e.g., moving velocity). Additionally, we ensure that the learned policy can effectively adhere to humanoid behavioral patterns by ensuring its distributional alignment with human motions. 
Note that this policy primarily focuses on ensuring task completion performance, without addressing safety issues that may arise from potential mismatches with the deployment environment.

The  {\it safety-recovery policy} aims at ensuring safety across various deployment environments. To address safety-critical scenarios during deployment, we construct an uncertainty set by designing an extreme-case estimation mechanism. Within these constructed scenarios, we propose a dynamic constraint guided by the Zero Moment Point (ZMP) to ensure stability under various disturbances.

While the goal-tracking and safety-recovery policies are primarily designed to enhance task performance and safety control, respectively, they lack a coordination mechanism to effectively control the trade-off between these objectives. To address this limitation, we propose a {\it high-level planning policy} that dynamically selects which policy to activate based on the scenario.

We conduct extensive experiments to evaluate the performance of HWC-Loco from four fundamental perspectives: 1) {\it effectiveness}, demonstrated across diverse terrains, 2) {\it robustness}, evaluated under varying scales of disturbances, 3) {\it naturalness}, exhibited in its ability to imitate human behavioral norms, 4) {\it scalability} evaluated in different robot embodiment and motion tracking tasks. These validations are performed in both simulated and real-world settings, demonstrating HWC-Loco as a foundational advancement for achieving reliable humanoid locomotion in safety-critical scenarios.


\vspace{-0.15in}
\section{Related Work}
\vspace{-0.1in}

\noindent\textbf{Learning-based Legged Robot Locomotion.} Various learning-based methods have been proposed for legged locomotion, such as quadrupedal locomotion \cite{rudin2022learning,shi2022reinforcement,chen2023learning,nahrendra2023dreamwaq,margolis2023walk,zhuang2023robot,he2024agile} and bipedal locomotion \cite{li2021reinforcement,kumar2022adapting,li2023robust,duan2024learning}. Despite their success, these approaches cannot be directly applied to humanoid robots because of the complex physical structure and the higher degrees of freedom. 
Recent advances in hardware and learning methods (e.g., RL algorithms) have enabled humanoid robots to move through diverse and complex environments. 
Some humanoid control policy navigates across diverse terrains using only proprioceptive information \cite{radosavovic2024real,gu2024humanoid,Gu2024Locomotion}. Others enhance performance by incorporating additional sensors, such as vision or LiDAR, to collect detailed environmental data, enabling robots to perform complex tasks like stair climbing and parkour jumps \cite{Zhuang2024Parkour,he2024PIM,sun2025learning}. While these approaches have demonstrated promising locomotion performance, they lack mechanisms to handle safety-critical scenarios that humanoid robots may encounter during real-world deployment.


\noindent\textbf{Whole-body control for humanoid robots.}
Learning effective whole-body control remains a central challenge for humanoid robotics. Prior works~\cite{luo2023perpetual,luo2024real,luo2024smplolympics} have demonstrated impressive results in simulated avatars, showcasing the feasibility of learning-based control strategies. More recently, studies such as~\cite{Cheng2024Exbody,he2024h2o,he2024omnih2o,fu2024humanplus,He2024HOVER,he2025asap,ze2025twist,li2025amo} have enabled realistic robots to imitate complex human-like motions, such as boxing and dancing by leveraging motion priors from human demonstrations. Additionally, previous methods employed multi-stage reward designs to accomplish complex tasks \cite{kim2024stage,Zhang2024wococo}. However, the success of these methods often relies on intricate reward shaping and struggles to generalize across different humanoid robots, particularly when motion priors are unavailable.


\vspace{-0.1in}
\section{Problem Formulation}\label{sec:Problem Formulation}
\vspace{-0.1in}

{\bf
Learning Environment.} 
We formulate the environment of the humanoid locomotion task with a 
Partially Observable Markov Decision Process (POMDP) $\mdp=(\mathcal{S}, \mathcal{A}, \mathcal{O}, P_\mathcal{T}, r,  \mu_0, \gamma)$, where:
1) Within the state space $\mathcal{S}$, a state $s\in\mathcal{S}$ records the environmental information and the robot's internal states. 
2) $\mathcal{A}$ denotes the action space, and action $a \in \mathcal{A}$ denotes the target joint angles that a PD controller uses to actuate the degrees of freedom (DOF). 
3) $o\in\mathcal{O}$ denotes an observation, which provides partial information about the state of the agent due to sensory limitations and environmental uncertainty. At a time step $t$, \(\boldsymbol{o}_{t}\) includes velocity command \( \boldsymbol{v}_t^{cmd} \in \mathbb{R}^{3} \) and proprioception \(\boldsymbol{o}_{t}^p\) that records a humanoid's internal state. Appendix \ref{Appendix:implementation} introduces the details.
By incorporating the temporal observations
\(\boldsymbol{o}_{t}^{H} = [\boldsymbol{o}_{t},\boldsymbol{o}_{t - 1},\cdots,\boldsymbol{o}_{t - H}]\), 
the control model can summarize a state as $s_t=[\boldsymbol{o}_{t}^H,\mathbb{P}_t]$, where $\mathbb{P}_t$ is the privileged information which the robot can't access in realistic deployment, including base velocity $v_t$, terrain height, external disturbance and ZMP features (see Section~\ref{subsec:high-level-policy}). 
4) $\transition\in \Delta^{\mathcal{S}}_{\mathcal{S}\times\mathcal{A}}$ denotes the transition function as a mapping from state-action pairs to a distribution of future states. 
5) $r$ denotes the reward functions, which typically consist of penalty, regularization, and task rewards. These reward signals significantly influence the optimality of the control policy, for which we provide a detailed introduction in the following.
6) $\mu_0\in\Delta^{\mathcal{S}}$ denotes the initial state distribution.
7) $\gamma\in(0,1]$ denotes the discounting factor. 
The robot's policy stops at a terminating state $\Tilde{s}$, and the corresponding terminating time is denoted as $T\in(0,\infty)$.

Under this POMDP, our goal is to learn a policy $\pi\in\Delta^{\mathcal{A}}_\mathcal{S}$ according to the objective as follows:
\begin{align}
    \max_\pi\mathcal{J}(\pi,\mdp)=\max_\pi\expect_{\mu_0,p_\mathcal{T},\pi}\left[\sum_{t=0}^{\infty}\gamma^t\left[\beta_\mathfrak{T}r_\mathfrak{T}(s_t,a_t)+\beta_\mathfrak{P}r_\mathrm{P}(s_t,a_t) +\beta_\mathfrak{R}r_\mathfrak{R}(s_t,a_t)\right]\right]\label{obj:rl}
\end{align}
In the task of humanoid locomotion, the total rewards can be generally represented as a weighted combination of task rewards $r_\mathfrak{T}$, penalty rewards $r_\mathrm{P}$, and regularization rewards $r_\mathfrak{R}$~\cite{he2024h2o,he2024omnih2o,Zhuang2024Parkour},
where the rewards have distinct semantic meanings:
\vspace{-0.1in}
\setlist[itemize]{leftmargin=*}
\begin{itemize}
\setlength\itemsep{0em}
\item {\it Task rewards} $r_\mathfrak{T}$ assess how effectively the agents achieve the goals of the current task, such as tracking velocities~\cite{Zhuang2024Parkour}, managing contacts~\cite{Zhang2024wococo}, and producing expressive motions~\cite{Cheng2024Exbody}. These task rewards are closely correlated with the optimality of the current control policy, making them an appropriate {\it objective to maximize} in this context.
\item{\it  Penalty rewards} $r_\mathrm{P}$ are used to prevent undesirable events. For example, in locomotion tasks, these penalties are applied when the humanoid falls 
or when the robot violates dynamic constraints, such as torque limits and joint position limits. To mitigate these negative outcomes, recent studies~\cite{he2024h2o,he2024omnih2o} have assigned a large weight \( \beta_\mathfrak{P} \) to these rewards. While these weights inherently function as Lagrange multipliers~\cite{Gu2024SafeRL}, they are predefined and not subject to optimization. Consequently, a more effective approach would be to model the control problem as a {\it constrained RL problem}.
\item{\it Regularization rewards} $r_\mathfrak{R}$  aligns the behavior of the humanoid with human preferences on motion styles or safe deployment. 
However, accurately specifying these regularizations remains challenging since they are often task-dependent, context-sensitive, and inherently rooted in human experts' experience~\cite{Chen2024LipschitzConstrainedPolicies}.
As a result, instead of manually specifying regularizations, we {\it align the robot's motion with human preferences} by imitating human behavior from motion datasets (e.g., AMASS dataset~\cite{Mahmood2019AMASS}), which contains rich signals about desirable behavior norms. 
\end{itemize}
\vspace{-0.05in}

\noindent\textbf{Constrained RL for Whole-Body Control.} Inspired by the aforementioned analysis, we formulate the whole-body control locomotion objective for humanoid robots as follows:
\begin{align}
    \max_\pi\expect_{\mu_0,p_\mathcal{T},\pi}\left[\sum_{t=0}^{\infty}\gamma^t r_\mathfrak{T}(s_t,a_t)\right]
    {s.t.}\; \mathcal{D}_{f}(\rho_{\mdp}^{\pi^{E}}\|\rho_{\mdp}^{\pi})\leq \epsilon_f\text{ and } \expect_{\tau\sim(\mu_0,p_\mathcal{T},\pi)}\left[\phi(\tau)\right] \geq \epsilon_\phi\label{obj:crl}
\end{align}
where 1) we replace the hard regularization in $r_\mathfrak{R}$ with a mimic learning objective $\mathcal{D}_{f}(\rho_{\mdp}^{\pi^{E}}\|\rho_{\mdp}^{\pi})$ that constrains the distributional divergence between the learned policy's occupancy measure $\rho_{\mdp}^{\pi}$ and the expert policy's occupancy measure $\rho_{\mdp}^{\pi^{E}}$. In this study we implement $\mathcal{D}_{f}$ by  Wasserstein distance (Section~\ref{subsec:low-level-Goal-policy}).
2) Instead of relying on the penalty reward $r_\mathrm{P}$, we use $\phi$ to capture the feasibility of the current trajectory generated by the policy $\pi$ under the environment $\mdp$.
Within the objective, only the task rewards 
$r_\mathfrak{T}$ are subject to maximization; other desired characteristics of humanoid robots are captured by constraints learned from human motion datasets. More importantly, this approach effectively eliminates the need for regularization on the humanoid robot's upper or lower pose, significantly enhancing the learning of whole-body locomotion policies. 

{\bf Robust Locomotion in Humanoid Robot.}
A common approach to learning humanoid locomotion policies involves training in a simulated environment before deploying the policies in a real-world setting. Due to the complexity of real-world environments, there is often a significant discrepancy between the training and deployment environments. In this study, we characterize this discrepancy by the mismatched POMDPs defined as follows: 
\begin{definition}
  (Mismatched POMDPs) We define a POMDP $\mdp:=(\cS,\cA,\transition, \mathcal{O}, r,\mu_0,\gamma)$, to be mismatched with another POMDP $\mdp^{\prime}:=(\cS,\cA,\transition^{\prime}, \mathcal{O}, r,\mu_0,\gamma)$ if they differ only in transition functions (i.e., $\transition\neq\transition^{\prime}$). We define a set of mismatched POMDPs as $\boldsymbol{\mathcal{\mdp}}_{m}=\{\mdp^{\transition},\mdp^{\transition^{\prime}},\dots\}$, where their transition functions differ from each other.
\end{definition}
Note that we follow previous studies~\cite{Moos2022RobustRL} and primarily focus on the mismatch in transition dynamics during learning and deployment. The uncertainty set of transition function~\cite{Viano2022RobustLearning} is represented as:
\begin{align}
    \trainsitionSet_{\alpha}^{L}=\{\alpha\transition^{L}+(1-\alpha)\bar{\transition},\;\forall{\bar{\transition}\in \trainsitionSet}\}\label{eqn:uncertainty-set}
\end{align}
where $\alpha$ specifies the scale of mismatch, $\transition^{L}$ denotes the transition function in the learning environment and $\trainsitionSet\subseteq\Delta^{\cS}_{\cS\times\cA}$ represents the set of all candidate transition functions.
While other factors, such as divergence in state-action spaces and initial state distribution $\mu_0$, can influence the performance during deployment, we find that the majority of factors affecting Sim2Real performance in humanoid locomotion tasks can be characterized by the mismatch in transition dynamics.
For example, if $\bar{\transition}$ denotes a Gaussian function, the underlying sensor noise can be captured by transitions in $\trainsitionSet_{\alpha}^{L}$. Additionally, if $\bar{\transition}$ denotes a projection from one spatial state to another, the variations of terrains in the deploying environment can be modeled by transitions in $\trainsitionSet_{\alpha}^{L}$~\cite{he2024PIM}.

To solve a robust RL problem, previous studies often consider a max-min objective $\max_\pi\min_{\mdp\in\boldsymbol{\mathcal{\mdp}}_{m}}\mathcal{J}(\pi,\mdp)$ where $\mathcal{J}(\pi,\mdp)$ denotes a standard RL objective~(\ref{obj:rl}). With this objective, while the agent can learn a conservative policy to ensure worst-case control performance, this policy often compromises the control performance, making the humanoid less effective at tracking the given commands. To address this issue, we focus on ensuring a worst-case feasibility constraint, thereby reducing the influence of mismatched POMDP on maximizing task rewards.
Accordingly, we update the CRL objective~(\ref{obj:crl}) to the robust humanoid locomotion objective represented as:
\vspace{-0.15in}
\begin{align}
\label{obj:robust-crl}\\[-0.8em]
    \max_\pi\min_{\textcolor{blue}{\widehat{\transition}}\in\trainsitionSet_{\alpha}^{L}}\expect_{\mu_0,\textcolor{red}{\transition^L},\pi}\left[\sum_{t=0}^{\infty}\gamma^t r_\mathfrak{T}(s_t,a_t)\right]
    {s.t.}\; \mathcal{D}_{f}(\rho_{\mdp^{\textcolor{red}{\transition^L}}}^{\pi^{E}}\|\rho_{\mdp^{\textcolor{red}{\transition^L}}}^{\pi})\leq \epsilon_f\text{ and } \expect_{\tau\sim(\mu_0,\textcolor{blue}{\widehat{\transition}},\pi)}\left[\phi(\tau)\right] \geq \epsilon_\phi\nonumber
\end{align}
where 1) $\trainsitionSet_{\alpha}^{L}$ represents the set of mismatched dynamics from the learning environment dynamics $\textcolor{red}{\transition^L}$ (defined in \ref{eqn:uncertainty-set}) and 2) $\rho_{\mdp}^{\pi}(s,a)=(1-\gamma)\pi(a|s)\sum_{t=0}^{\infty}\gamma^{t}p(s_{t}=s|\pi,\mdp)$ defines the normalized occupancy measure of policy $\pi$ under the environment $\mdp$. 
Intuitively, we aim to learn a robust control policy that ensures feasibility across all mismatched transition functions $\textcolor{blue}{\widehat{\transition}}\in\trainsitionSet_{\alpha}^{L}$, 
thereby guaranteeing the safety of policy $\pi$. However, for the reward-maximizing and mimic learning objectives, we optimize the policy based on the transition dynamics in the learning environment $\textcolor{red}{\transition^L}$, rather than concentrating solely on worst-case guarantees.

\vspace{-0.1in}
\section{Hierarchical Whole-Body Control for Humanoid Robust Locomotion}
\label{sec:HWC-Loco_approach}
\vspace{-0.1in}
\begin{figure*}[htbp]
    \centering
    \includegraphics[width=1.0\linewidth]{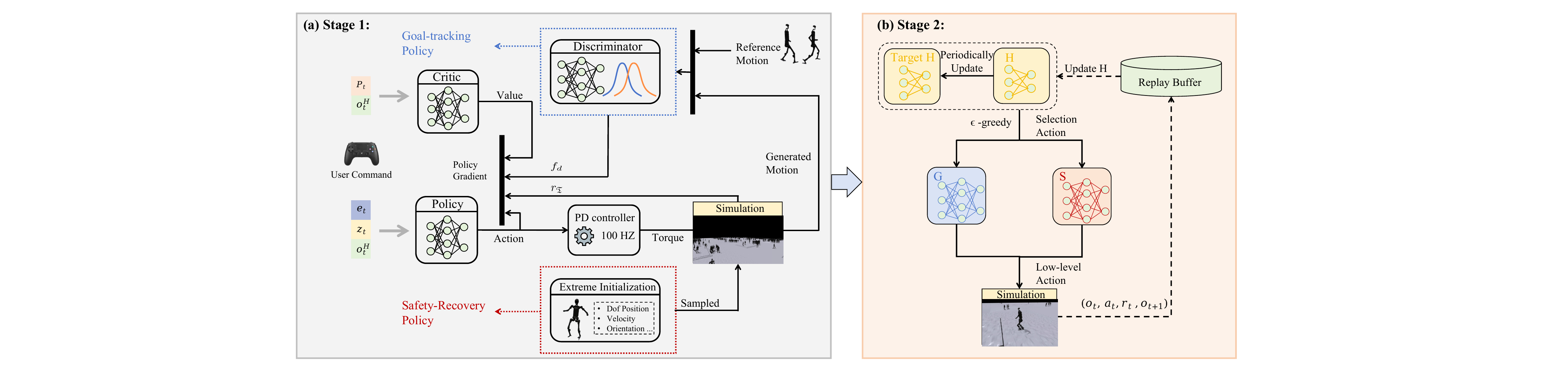}
    \caption{ Overview of HWC-Loco: The framework consists of two stages: (a) Training {\it goal-tracking policy} to effectively enable human-like locomotion across diverse terrains (Section~\ref{subsec:low-level-Goal-policy}) and {\it safety recovery policy} to recover from safety-ciritical states (i.e., extreme-case) (Section~\ref{subsec:low-level-Recovery-policy}).
    (b) Training the {\it high-level planning policy} to select between the two pre-trained low-level policies (Section~\ref{subsec:high-level-policy}), 
    thereby ensuring locomotion stability and consistency.} 
    \label{fig:Framework}
    \vspace{-0.2in}
\end{figure*}

In practice, developing an end-to-end solver for the robust locomotion problem (\ref{obj:robust-crl}) is challenging, as it involves multiple deployment environments and constraints that must be adhered to. To tackle this issue, we propose dividing the objective into two stages: goal-tracking and safety recovery. The goal-tracking policy maximizes rewards while mimicking human behaviors (see Section~\ref{subsec:low-level-Goal-policy}). When mismatched environmental dynamics pose a risk to the safety of the control policy, HWC-Loco switches to the safety recovery policy, which manages safety-critical events to ensure the feasibility of humanoid control (see Section~\ref{subsec:low-level-Recovery-policy}). To dynamically identify the "sweet spot" for policy switching or activation, we introduce a high-level policy that coordinates the low-level policies based on historical observations and the robot's current status (see Section~\ref{subsec:high-level-policy}). Figure~\ref{fig:Framework} illustrates the training pipeline of our HWC-Loco.

\vspace{-0.1in}
\subsection{Learning Goal-Tracking Policy for Efficient Humanoid Locomotion}\label{subsec:low-level-Goal-policy}
\vspace{-0.05in}
The goal-tracking policy primarily focuses on tracking the provided command within the learning environment efficiently and naturally. This is achieved by optimizing the following objective function:
\vspace{-0.15in}
\begin{align}
    &\max_{\pi_1}\expect_{\rho_{\mdp^{\textcolor{red}{\transition^L}}}^{\pi_1}}\left[\sum_{t=0}^{\infty}\gamma^t r_\mathfrak{T}(s_t,a_t)\right]
    {s.t.}\; \mathcal{D}_{f}(\rho_{\mdp^{\textcolor{red}{\transition^L}}}^{\pi^{E}}\|\rho_{\mdp^{\textcolor{red}{\transition^L}}}^{\pi_1})\leq \epsilon_f \label{obj:goal-tracking}
\end{align}
Note that compared to the complete objective
(\ref{obj:robust-crl}), this goal-tracking objective 
focuses only on maximizing performance in the training environments' dynamics \textcolor{red}{$\transition^L$}.
Within this objective, the scale of the cumulative task rewards, $\sum_{t=0}^{\infty} \gamma^tr_\mathfrak{T}$ reflects how effectively the policy $\pi$ tracks locomotion-related commands.
Appendix \ref{Appendix:implementation} specifies the configuration of our task reward.


In addition to the task rewards, a crucial aspect of the goal-tracking objective (\ref{obj:goal-tracking}) is evaluating the distance between the occupancy measures of the trained policy and the expert policy 
In the following, striving for concise representation, we represent $\rho_{\mdp^{\textcolor{red}{\transition^L}}}^{\pi^{E}}$ and $\rho_{\mdp^{\textcolor{red}{\transition^L}}}^{\pi}$ by $\rho^{\pi^{E}}$ and $\rho^{\pi}$. Driven by information theory, previous works~\cite{Hussein2017Imitation} primarily implement $\mathcal{D}_{f}$ using metrics such as Kullback-Leibler (KL) divergence or Jensen-Shannon (JS) divergence. However, these metrics fail when the support of $\rho^{\pi^{E}}$ and $\rho^{\pi}$ has limited or zero overlap. In this study, we consider implement $\mathcal{D}_{f}$ with Wasserstein-1 distance under the Kantorovich-Rubinstein duality~\cite{villani2009optimal} such that:
\begin{align}
    \mathcal{D}_{f}(\rho^{\pi^{E}}\|\rho^{\pi_1}) = \sup_{\|f_d\|_{L} \leq 1} \mathbb{E}_{x \sim \rho^{\pi^{E}}}[f_d(x)] - \mathbb{E}_{x \sim \rho^{\pi_1}}[f_d(x)]\label{eqn:f-divergence}
\end{align}
where the bounded Lipschitz-norm $\|f_d\|_{L} \leq 1$ ensure the smoothness of functions $f_d$. Inspired by adversarial imitation learning~\cite{Ho2016GAIL,Zhang2020fGAIL}, an effective and intuitive implementation of $f_d$ is a discriminator for measuring whether $x$ is generated by human experts, such that $f_d: \mathcal{X} \rightarrow \mathbb{R}$. It can be learned by maximizing the following objective:
\begin{equation}
    \max_{f_d} \; \mathbb{E}_{x \sim \rho^{\pi^{E}}}[f_d(x)] - \mathbb{E}_{x \sim \rho^{\pi_1}}[f_d(x)] + \beta \cdot \mathbb{E}_{\hat{x} \sim [\rho^{\pi^{E}}, \rho^{\pi_1}]} \left[ (\|\nabla_{\hat{x}} f_d(\hat{x})\|_2 - 1)^2 \right]
    \label{obj:discrtiminator}
\end{equation}


where weighted by $\beta$, the gradient penalty term enforces the Lipschitz continuity required by the Wasserstein distance (i.e., $\|f_d\|_{L} \leq 1$). 
During training, $\rho^{\pi^{E}}$ represents the density of expert demonstration dataset $\dataset^E$, which is built from the CMU MoCap dataset \cite{cmudataset}.
Additionally, we include data about humans standing, walking, and running to learn various human behaviors. We then retarget the motion data into a humanoid robot-compatible format, which we use as the training dataset.
By substituting $\mathcal{D}_f$ with equation (\ref{eqn:f-divergence}), the goal-tracking objective~(\ref{obj:goal-tracking}) can be simplified as:
\begin{equation}
\max_{\pi_1} \mathbb{E}_{\rho^{\pi_1}}\left[\sum_{t=0}^{\infty} \gamma^t r_\mathfrak{T}(s_t,a_t)\right] 
\quad \text{s.t.} \quad \mathbb{E}_{\rho^{\pi^E}}[f_d(s^d)] - \mathbb{E}_{\rho^{\pi_1}}[f_d(s^d)] \leq \epsilon_f
\label{obj:goal-tracking-W1}
\end{equation}
where $s^d$ records the necessary state information for the discriminator.
(See details in Appendix \ref{Appendix:Discriminator}).
Given that this formulation of constrained RL problem has a zero duality gap \cite{Paternain2019CRL}, we can transform it into an unconstrained objective by analyzing the Lagrange dual form of the original constrained RL problem:
\vspace{-0.05in}
\begin{align}
     \max_{\pi_1}\expect_{\rho^{\pi_1}}\left[\sum_{t=0}^{\infty}\gamma^t \left[r_\mathfrak{T}(s_t,a_t)-\lambda f_d(s^d)\right]\right]\label{obj:goal-tracking-unconstrained}
\end{align}
where $\lambda$ denotes the optimal Lagrange multiplier. In practical implementation, the policy $\pi$ and discriminator $f_d$ can intuitively represent the generator and the discriminator under the adversarial learning framework.
In implementation, we utilize the proximal policy optimization algorithm~\cite{schulman2017proximal} to update the goal-tracking policy. By alternatively updating (\ref{obj:discrtiminator}) and (\ref{obj:goal-tracking-unconstrained}), we can develop a goal-tracking policy that effectively maximizes task rewards while closely mimicking the expert's behavior. 
\vspace{-0.1in}
\subsection{Learning Safety Recovery Policy for Handling Safe-Critical Events}\label{subsec:low-level-Recovery-policy}
\vspace{-0.05in}

Our safety recovery policy primarily focuses on handling emergency events, thereby recovering the robot from safety-critical situations and preventing failures of locomotion tasks, such as loss of balance or control policy malfunctions. To learn the safety-recover policy, we mainly study the complete objective~(\ref{obj:robust-crl}). Similarly, as it is introduced in Section~\ref{subsec:low-level-Goal-policy}, we implement the divergence metric $\mathcal{D}_f$  with Wasserstein-1 distance, deriving the following objective:
\vspace{-0.05in}
\begin{equation}
\max_{\pi_2}\min_{\textcolor{blue}{\widehat{\transition}}\in\trainsitionSet_{\alpha}^{L}}\expect_{\mu_0,\textcolor{blue}{\widehat{\transition}},\pi_2}\left[\sum_{t=0}^{\infty}\gamma^t r_\mathfrak{T}(s_t,a_t)-\lambda f_d(s_t,a_t)\right] \;\;{s.t.}\;\; \expect_{\tau\sim(\mu_0,\textcolor{blue}{\widehat{\transition}},\pi_2)}\left[\phi(\tau)\right] \geq \epsilon_\phi\label{obj:safety-recover}
\end{equation}
\vspace{-0.10in}

where $\epsilon_\phi$ is a hyperparameter. By comparing this objective with the goal-tracking objective~(\ref{obj:goal-tracking-unconstrained}), while both consider task rewards and human mimicking performance, their key differences lie in the incorporation of uncertainty in environmental dynamics and feasibility constraints in policy updates. These additions necessitate the construction of an extreme-case uncertainty set and the modeling of safety constraints by learning the feasibility function $\phi(\cdot)$.

\vspace{-0.05in}
\textbf{Extreme-case Uncertainty Set.} 
Real-world deployment has complex terrains, external disturbances, hardware malfunctions, and sensor noise.
Those factors may lead to extreme cases like falling or unexpected behaviors. 
To model them in the simulation (i.e., modelling $\textcolor{blue}{\widehat{\transition}}$s), 
We: 1) apply multi-scale external forces and torques to the humanoid body, 2) introduce random, high-intensity noise to the proprioceptive information and PD gains, 3) resample velocity commands to simulate malicious velocity inputs~\cite{shi2024rethinking} within the goal-tracking policy, and 4) apply domain randomization to the training environment. Together, these dynamic variations construct the uncertainty set $\trainsitionSet_{\alpha}^{L}$, under which we learn the safety-recover policy. Appendix \ref{Appendix:Extreme Initialization} shows our implementation details.

\vspace{-0.05in}
\textbf{Zero-Moment Point (ZMP) Constraint.} 
A bipedal robot can be generally modeled by a linear inverted pendulum~\cite{morisawa2006biped,harada2006analytical}, where ZMP refers to the point where the ground reaction force has no horizontal moment. If the ZMP exits the support polygon (typically the foot’s contact area), the robot will quickly lose balance~\cite{wieber2006trajectory,feng2016robust,scianca2020mpc,xie2025humanoid}. Stability is governed by gravity and the Center of Mass (CoM) accelerations, which also account for disturbances like obstacles and slippery surfaces. Under this assumption, we implement the feasibility indicator in the safety recovery objective~(\ref{obj:safety-recover}) as follows:
\vspace{-0.05in}
\begin{align}
\phi(s,a) = \|\mathbf{p}_\text{ZMP} (s,a) - \mathbf{p_{\text{ac}}}\|_2 \;\;\text{where}\;\; \mathbf{p}_{\text{ZMP}}(s,a) = \mathbf{p}_{\text{CoM}}(s,a) - \frac{z_{\text{CoM}}(s,a)}{g} \cdot \ddot{\mathbf{p}}_{\text{CoM}}(s,a)
\end{align}
where 1) the vector $\mathbf{p}_{\text{CoM}}$ denotes the $x$ and $y$-coordinates of CoM position. 2) $z_{\text{CoM}}$ is the height of the CoM. 3) $g$ represents the acceleration due to gravity. 4) $\ddot{\mathbf{p}}_{\text{CoM}}$ indicates the accelerations of the CoM in the $x$ and $y$-directions. 5) \(\mathbf{p}_{\text{ac}}\) represents the center of the support polygon (specific to each robot and known in advance). 
Intuitively, $\phi$ provides real-time insights into the robot's stability and varies in its representation depending on the current support phase. Therefore, the humanoid robot can assess its stability in real time and use whole-body coordination to satisfy the ZMP constraint.

\vspace{-0.1in}
\subsection{High-Level Planning for Policy Transition}\label{subsec:high-level-policy}
\vspace{-0.05in}

\begin{wrapfigure}{r}{0.45\textwidth} 
  \centering
  \vspace{-30pt}  
  \includegraphics[width=0.95\linewidth]{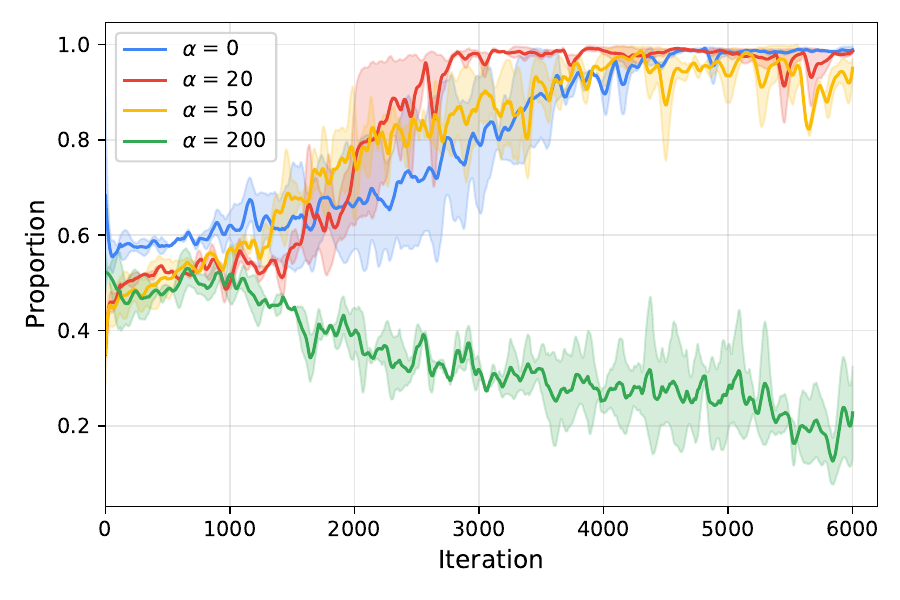} 
  \vspace{-8pt}
  \caption{Proportion of goal-tracking policy.}
  \label{fig:Proportion}
  \vspace{-5pt}
\end{wrapfigure}
While the low-level policies can achieve varying levels of optimality,
a fundamental challenge is coordinating these policies to ensure that a humanoid robot can efficiently complete tasks while adhering to safety and robustness requirements. We introduce a high-level planning policy that dynamically determines which policy to activate based on the specific state of humanoid robots. 
The task of policy selection has a discrete action space, and we learn this high-level policy $\pi_0$ based on the following objective: 
\vspace{-0.15in}
\begin{align}
   \hspace{-0.15in}\max_{\pi_0}\expect\left[ \sum_{t=0}^\infty\gamma^t\left[r_\mathfrak{T}(s_t,\bar{a}_t) -\mathds{1}(\bar{a}_{t-1} \neq \bar{a}_{\text{t}}) - \alpha \mathds{1}(s_t)\right]\right]\label{obj:high-level-policy}
\end{align}
where 1) the policy selection action $\bar{a}_t\in\{0,1\}^2$ denotes a one-hot vector featuring whether the goal-tracking policy $\pi_1$ and safety-recovery policy $\pi_2$, 2) $\mathds{1}(\bar{a}_{t-1} \neq \bar{a}_{\text{t}})$ denotes the continuality identifier for discouraging the over-frequent switch of policies, and  3) $\mathds{1}(s_t)$ denotes the termination identifier for preventing the failing of locomotion tasks. (2) and (3) are both $r_\mathrm{P}$ defined in Section \ref{sec:Problem Formulation}. More importantly, within this objective, by adjusting $\alpha$, practitioners can adjust the trade-off safety guarantee and task completion. Training and algorithmic details can be found in Appendix \ref{Appendix:implementation}.

We evaluate the average activation proportion of the goal-tracking policy during training.
As shown in Figure~\ref{fig:Proportion}, when $\alpha = 0$, 20, or 50, the goal-tracking policy $\pi_1$ is predominantly selected, with the recovery policy $\pi_2$ triggered only in rare dangerous cases. In contrast, when $\alpha = 200$, the sparsity of termination rewards induces training instability, resulting in overly conservative behavior.

\textbf{Practical Deployment.} The training of humanoid control policy commonly relies on the privileged information $\mathbb{P}_t$, such as external disturbances, terrain dynamics, etc. We follow ~\cite{nahrendra2023dreamwaq} and infer $\mathbb{P}_t$ by training a VAE-based estimator $P(e_t, z_t|o^H_t)$ where $z_t$ captures the contextual embedding of the $\mathbb{P}_t$, and $e_t$ consists of body velocity and ZMP features. To enhance the representation of $\phi(\cdot)$, we introduce frequency encoding, similar to the approach in \cite{mildenhall2021nerf}, to capture the subtle variations in $\phi(\cdot)$ for the policy. Additionally, instead of the zero-mean Gaussian prior, we introduce a learnable prior similar to \cite{luo2023universal}. Moreover, we include a certain degree of domain randomization in the learning of HWC-Loco policies. The details of domain randomization are in Appendix \ref{Appendix:Extreme Initialization}.
\vspace{-0.1in}
\section{Experiment}
\label{Experiment}
\vspace{-0.1in}
\textbf{Experiment Settings.}
To conduct a comprehensive evaluation, we quantify the control performance of HWC-Loco in simulated (Issac Gym~\cite{makoviychuk2021isaac}) and realistic environments from the following perspectives:
1) {\it Effectiveness}: How effectively does HWC-Loco navigate across diverse terrains?
2) {\it Robustness}: How well can HWC-Loco stabilize the humanoid robot under varying levels of disturbance?
3) {\it Naturalness}: How well does HWC-Loco imitate human-like movement norms?
4) {\it Scalability}: How effectively does HWC-Loco generalize to diverse embodiments and tasks? 

Our evaluation metric includes a) {\it Success Rate}: The proportion of successful traverse across different scenarios. \cite{cuiadapting}
b) {\it Goal Tracking} performance: The ability to accurately follow velocity commands by maximizing task rewards $r_\mathfrak{T}$  detailed in Appendix \ref{Appendix:implementation} \cite{Chen2024LipschitzConstrainedPolicies}.
c) {\it Human-Like} behavior: Measured as the Wasserstein-1 distance between the robot's and sampled human motions. 

\textbf{Comparison Methods.}
To evaluate the effectiveness of different components in HWC-Loco, we design a comparison method using an ablation approach as follows:
1) {\it HWC-Loco-l} sets $\alpha$ to a lower value, thereby reducing the sensitivity of the high-level policy to failure events.
2) {\it Goal-Tracking Policy} removes the safety recovery mechanism and relies solely on the goal-tracking policy $\pi_1$ (Section~\ref{subsec:low-level-Goal-policy}).
3) {\it DreamWaQ-Humanoid} further removes the human imitation objective $\mathcal{D}_{f}$ from objective~(\ref{obj:goal-tracking}), effectively reducing our method to an adaptation of DreamWaQ \cite{nahrendra2023dreamwaq} for humanoid control.
4) To highlight the benefits of our hierarchical policy, we compare a recent advancement of DreamWaQ, called {\it AHL} \cite{cuiadapting}, which employs two-phase training for updating the locomotion policy. All policies are trained using three random seeds and evaluated in 1000 distinct environments. 

\vspace{-0.05in}
\subsection{Effectiveness: Locomotion across Diverse Terrains}\label{subsec:exp-Scalability}
\vspace{-0.05in}

\textbf{Effectiveness in Simulated Environments.} 
We evaluate the policies on challenging terrains, specifically slopes and stairs. To enable a comprehensive comparison, evaluations are performed under both low-speed and high-speed command conditions. Low-speed commands are sampled as $v_{\text{x}}^c \sim [0.0, 1.0]$, $v_{\text{y}}^c \sim [-0.3, 0.3]$, and $w_{\text{z}}^c \sim [-0.5, 0.5]$, while high-speed commands follow $v_{\text{x}}^c \sim [1.0, 2.0]$, $v_{\text{y}}^c \sim [-0.6, 0.6]$, and $w_{\text{z}}^c \sim [-1.0, 1.0]$. Here, $v_{\text{x}}^c$ and $v_{\text{y}}^c$ denote the commanded linear velocities along the x- and y-axes, respectively, and $w_{\text{z}}^c$ denotes the commanded angular velocity around the z-axis. Table \ref{tab:comparision} shows that our HWC-Loco achieves the highest success rate across all evaluated scenarios, while maintaining promising goal tracking and human mimic performance. Notably, in the high-speed stair terrain setting, HWC-Loco has a significantly higher success rate than all other policies, although its goal-tracking ability slightly decreases. This indicates that the policy prioritizes stability over aggressive goal pursuit. For example, when given a high-speed command, HWC-Loco enables the policy to adapt dynamically in safety-critical situations (e.g., navigating stairs mid-air) rather than rigidly maintaining high velocity. Comparably, when downplaying the sensitivity to safety-critical events and removing the safety-recovery policy, the success data drops significantly from nearly 85\% to around 60\% in the testing environment with stairs and high-speed commands.

\textbf{Effectiveness in Realistic Environments.} 
Figure~\ref{fig:Climb_stairs} (see Appendix~\ref{Appendix:Deployment}) demonstrates the robot's ability to climb 15\,cm stairs and 20° slopes. When faced with challenging terrain, the robot prioritizes safety by activating the recovery policy. After regaining stability, it smoothly transitions back to goal tracking, showcasing its adaptive control in dynamic and uncertain environments. Figure~\ref{fig:outdoor_results} (see Appendix~\ref{Appendix:Deployment}) further illustrates the robot’s effectiveness in real-world outdoor settings, where it successfully navigates across diverse terrains, including flat ground, grassy surfaces, and slopes.


\begin{table*}[h!]
\vspace{-0.1in}
    \centering
    \caption{Locomotion performance in simulated environments. Each evaluation runs for 1200 steps, which is equivalent to 12 seconds of real clock time.}
    \label{tab:comparision}
    \small
    \setlength{\tabcolsep}{4pt} 
    \resizebox{1.0\textwidth}{!}{
    \begin{tabular}{lcccccc}
        \toprule
        \textbf{Method} & \multicolumn{3}{c}{\textbf{Slopes}} & \multicolumn{3}{c}{\textbf{Stairs}} \\
        \cmidrule(lr){2 - 4} \cmidrule(lr){5 - 7}
         & \textbf{Success Rate(\%) $\uparrow$} & \textbf{Goal Tracking $\uparrow$} & \textbf{Human - like $\downarrow$} 
         & \textbf{Success Rate(\%) $\uparrow$} & \textbf{Goal Tracking $\uparrow$} & \textbf{Human - like $\downarrow$} \\
        \midrule
        \textbf{(a) Low Speed} & & & & & & \\
        DreamWaQ       & $92.31 \pm 0.40$ & $1.19 \pm 0.02$ & $3.52 \pm 0.05$ 
                          & $74.32 \pm 1.30$ & $1.19 \pm 0.03$ & $3.53 \pm 0.02$ \\
        AHL           & $98.83 \pm 0.15$ & $1.21 \pm 0.01$ & $3.41 \pm 0.04$ 
                          & $93.73 \pm 0.70$ & $1.21 \pm 0.04$ & $3.44 \pm 0.01$ \\
        Goal-tracking Policy        & $99.90 \pm 0.04$ & $\boldsymbol{1.31 \pm 0.00}$ & $\boldsymbol{3.30 \pm 0.05}$ 
                          & $96.60 \pm 0.14$ & $\boldsymbol{1.31 \pm 0.00}$ & $\boldsymbol{3.31 \pm 0.01}$ \\
        HWC-Loco-l & $\boldsymbol{100.00 \pm 0.00}$ & $1.23 \pm 0.02$ & $3.32 \pm 0.03$ 
                          & $99.80 \pm 0.08$ & $1.21 \pm 0.02$ & $3.31 \pm 0.06$ \\
        HWC-Loco & $\boldsymbol{100.00 \pm 0.00}$ & $1.22 \pm 0.02$ & $3.32 \pm 0.05$ 
                          & $\boldsymbol{99.98 \pm 0.01}$ & $1.19 \pm 0.03$ & $3.34 \pm 0.03$ \\
        \midrule
        \textbf{(b) High Speed} & & & & & & \\
        DreamWaQ       & $90.46 \pm 0.43$ & $1.05 \pm 0.01$ & $3.63 \pm 0.04$ 
                          & $60.58 \pm 0.64$ & $1.06 \pm 0.01$ & $3.66 \pm 0.02$ \\
        AHL           & $97.36 \pm 0.23$ & $1.12 \pm 0.01$ & $3.76 \pm 0.02$ 
                          & $67.48 \pm 0.77$ & $1.09 \pm 0.01$ & $3.68 \pm 0.04$ \\
        Goal-tracking Policy        & $98.51 \pm 0.45$ & $\boldsymbol{1.13 \pm 0.00}$ & $3.52 \pm 0.03$ 
                          & $72.60 \pm 0.97$ & $\boldsymbol{1.11 \pm 0.00}$ & $3.43 \pm 0.02$ \\
        HWC-Loco-l & $99.95 \pm 0.02$ & $1.12 \pm 0.02$ & $\boldsymbol{3.43 \pm 0.04}$ 
                          & $78.92 \pm 0.45$ & $1.10 \pm 0.01$ & $\boldsymbol{3.39 \pm 0.05}$ \\
        HWC-Loco & $\boldsymbol{100.0 \pm 0.00}$ & $1.12 \pm 0.02$ & $3.44 \pm 0.09$ 
                          & $\boldsymbol{84.34 \pm 0.43}$ & $1.07 \pm 0.03$ & $3.46 \pm 0.03$ \\
        \bottomrule
    \end{tabular}
    }
\vspace{-0.2in}
\end{table*}



\subsection{Robustness: Stable Control under Disturbances}\label{subsec:exp-Disturbance}
\vspace{-0.1in}
\textbf{Robustness under Simulated Disturbances.} To demonstrate the robustness of HWC-Loco, we conduct extensive disturbance tests in simulation. The robots are commanded to follow randomly sampled velocities within the training distribution while navigating a uniformly mixed terrain comprising flat ground, obstacles, slopes, and stairs. Under these general locomotion settings, we design three types of disturbances as follows: 1) External force/torque disturbances, where random forces and torques (up to 200~N and 200~N$\cdot$m) are applied to each robot link, either at a low frequency of 1~Hz or as constant perturbations lasting 0.5 seconds, following prior setups~\cite{weng2023comparability,lee2020learning}. 2) Impulse disturbances on the CoM, implemented by directly altering the robot’s CoM velocity~\cite{weng2023comparability,gu2024humanoid}. We simulate both low-impulse and high-impulse scenarios, with linear velocity perturbations of 0--1~m/s or 1--2~m/s, and angular velocity perturbations of 0--0.5~rad/s or 0.5--1.0~rad/s, respectively. 3) Payload disturbances, where an additional mass uniformly sampled from either 0 to 5~kg or 0 to 10~kg is added to upper-body links~\cite{margolis2023walk,radosavovic2024real}, along with internal perturbations from sensor and actuator noise (see Table~\ref{tab:obs_noise} in Appendix~\ref{Appendix:Extreme Initialization}), to simulate unmodeled dynamics and varying mass distribution scenarios.

As shown in Table~\ref{tab:robustness_test}, HWC-Loco consistently achieves the highest success rates across all types of disturbances. Specifically, under constant disturbances, HWC-Loco demonstrates superior performance with a success rate of 75.95\%, which is nearly 25\% higher than that of DreamWaQ. It validates the effectiveness of HWC-Loco’s recovery mechanism in resisting strong and persistent forces. Under low-impulse disturbances, all methods maintain relatively high success rates, with HWC-Loco again achieving the best performance at 94.84\%. When facing more challenging high-impulse disturbances, the performance gap becomes more pronounced. HWC-Loco reaches a success rate of 81.27\%, outperforming all baselines by a significant margin. This result indicates that the recovery mechanism effectively mitigates the impact of sudden impulses by enabling the robot to quickly lower its body and wave its arms to regain stability. In the payload test, HWC-Loco also demonstrates the best performance. Although it s not explicitly trained on scenarios such as adding 10~kg to the hands, it maintains a high success rate. This highlights the robustness of HWC-Loco to unseen disturbances.

\begin{table}[htbp]
\vspace{-0.2in}
\centering
\caption{Robustness of Locomotion across Various Terrains under Different Disturbances}
\resizebox{\linewidth}{!}{
\begin{tabular}{lcc|cc|cc}
\toprule
\textbf{Policy} 
& \multicolumn{2}{c|}{\textbf{External Force/Torque Disturbances}} 
& \multicolumn{2}{c|}{\textbf{Impulse Disturbances on CoM}} 
& \multicolumn{2}{c}{\textbf{Payload on Upper Body}} \\
\cmidrule(lr){2-3} \cmidrule(lr){4-5} \cmidrule(lr){6-7}
& \textbf{Low-freq. $\uparrow$} & \textbf{Constant $\uparrow$} 
& \textbf{Low-imp. $\uparrow$} & \textbf{High-imp. $\uparrow$} 
& \textbf{Low Payload $\uparrow$} & \textbf{High Payload $\uparrow$} \\
\midrule
DreamWaQ      
& $85.92 \pm 1.64$ & $51.31 \pm 0.97$ 
& $85.24 \pm 0.37$ & $45.34 \pm 0.41$ 
& $67.63 \pm 3.10$ & $55.04 \pm 2.82$ \\
AHL           
& $87.15 \pm 2.86$ & $60.72 \pm 0.36$ 
& $85.87 \pm 1.47$ & $62.94 \pm 1.29$ 
& $79.29 \pm 0.65$ & $59.16 \pm 1.60$ \\
Goal-tracking Policy
& $90.00 \pm 2.20$ & $61.20 \pm 0.61$ 
& $88.90 \pm 0.60$ & $58.13 \pm 0.78$ 
& $78.34 \pm 1.23$ & $61.44 \pm 1.00$ \\
HWC-Loco-l    
& $92.69 \pm 0.56$ & $68.60 \pm 0.28$ 
& $92.79 \pm 1.75$ & $77.09 \pm 0.79$ 
& $84.11 \pm 2.66$ & $63.96 \pm 0.09$ \\
HWC-Loco      
& $\boldsymbol{95.88 \pm 0.37}$ & $\boldsymbol{75.95 \pm 0.66}$ 
& $\boldsymbol{94.84 \pm 0.54}$ & $\boldsymbol{81.27 \pm 0.80}$ 
& $\boldsymbol{87.43 \pm 0.92}$ & $\boldsymbol{69.86 \pm 1.00}$ \\
\bottomrule
\end{tabular}
}
\label{tab:robustness_test}
\vspace{-0.2in}
\end{table}

\textbf{Robustness in Realistic Deployment.} We deploy HWC-Loco on a real humanoid robot and evaluate its performance under various external force disturbances. Figure~\ref{fig:Disturbance} (see Appendix~\ref{Appendix:Deployment}) shows policy switching and associated changes in roll and pitch angles under continuous perturbations, including pushes, pulls, and kicks. 
Upon encountering disturbances, the robot promptly switches to the recovery policy, adjusting its posture and gait to regain stability. Importantly, the controller does not rely solely on recovery mode but dynamically switches between goal-tracking and recovery policies, thereby adapting the action distribution to environmental changes. Figure~\ref{fig:Outdoor_disturbance} (see Appendix~\ref{Appendix:Deployment}) further demonstrates the robot’s robustness to disturbances in outdoor settings.

\vspace{-0.1in}
\subsection{Naturalness: Imitating Human-like Movements}
\label{subsec:exp-Naturalness}
\vspace{-0.1in}
This experiment assesses the robot’s ability to imitate human-like locomotion under varying velocity commands. As shown in Figure~\ref{fig:NaturalWalking}, the robot remains stationary at 0 m/s, and exhibits natural walking behavior at 1.0 m/s, characterized by straight elbows and coordinated arm swings. At 2.5 m/s, it transitions to running, with longer strides and increasingly bent elbows, closely resembling human running. As the velocity increases, both stride length and step frequency adjust dynamically, mimicking typical human locomotion. In contrast, prior approaches \cite{gu2024humanoid, gu2024advancing, cuiadapting, zhang2024whole} often maintain a fixed step frequency regardless of the commanded velocity. Some even exhibit in-place stepping at 0,m/s \cite{zhang2024whole, cuiadapting}, resulting in unnatural behavior for tasks requiring a steady stance, such as manipulation.




\vspace{-0.1in}
\subsection{Scalability: Extending HWC-Loco to New Tasks}
\vspace{-0.05in}
\textbf{Cross Embodiment Validation.} We deploy HWC-Loco on the Unitree G1 humanoid robot to evaluate its generalization capability under the same locomotion settings. As shown in Table~\ref{tab:Embodiments}, G1 outperforms H1 across all metrics. A detailed analysis is provided in Appendix~\ref{Appendix:implementation}.

\vspace{-0.05in}
\begin{figure}[htbp]
\centering
\begin{minipage}[t]{0.35\linewidth}
    \centering
    \includegraphics[width=\linewidth]{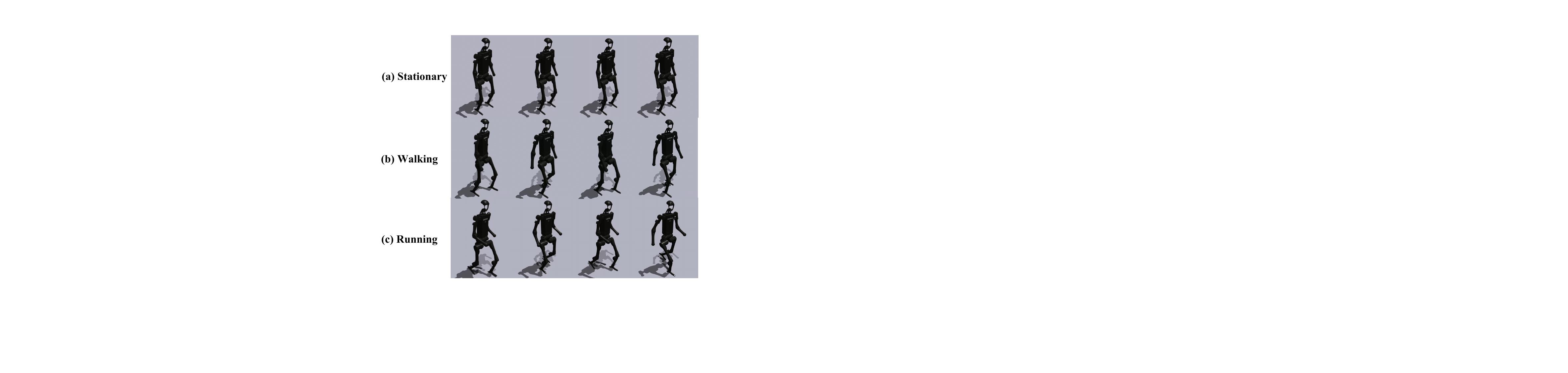}
    \vspace{-0.2in}
    \caption{Human-like behavior.}
    \label{fig:NaturalWalking}
\end{minipage}
\hfill
\begin{minipage}[t]{0.55\linewidth}
    \vspace{-1.5in}
    \centering
    \captionof{table}{Comparison of Embodiments}
    \label{tab:Embodiments}
    \resizebox{\linewidth}{!}{
    \begin{tabular}{lccc}
    \toprule
    \textbf{} & \textbf{Success Rate $\uparrow$} & \textbf{Goal Tracking $\uparrow$} & \textbf{Human-like $\downarrow$} \\
    \midrule
    \textbf{Unitree H1} & $97.13\pm 0.43$ & $1.10\pm0.00$ & $3.18 \pm0.01$ \\
    \textbf{Unitree G1} & $\boldsymbol{98.14\pm 0.35}$ & $\boldsymbol{1.12\pm 0.01}$ & $\boldsymbol{3.11\pm 0.03}$ \\
    \bottomrule
    \end{tabular}
    }
    \vspace{1em}
    
    \captionof{table}{Motion Tracking under Disturbances}
    \label{tab:ExpressiveMotion}
    \resizebox{\linewidth}{!}{
    \begin{tabular}{lccc}
    \toprule
    \textbf{} & \textbf{Punching} $\uparrow$ & \textbf{Dancing} $\uparrow$ & \textbf{Expressive Walking} $\uparrow$ \\
    \midrule
    \textbf{HWC-Loco} & $\boldsymbol{94.01\pm 0.49}$ & $\boldsymbol{86.44\pm0.60}$ & $\boldsymbol{94.53\pm0.23}$ \\
    \textbf{Motion-tracking} & $90.85\pm 0.74$ & $81.64\pm 0.64$ & $89.63\pm 0.27$ \\
    \bottomrule
    \end{tabular}
    }
\end{minipage}
\vspace{-0.1in}
\end{figure}



\textbf{Expressive Motion Tracking} 
\label{Appendix:Expressive Motion}
We follow the approach of ExBody\cite{Cheng2024Exbody,ji2024exbody2} and set expressive motion as the goal for HWC-Loco. We train the model following the same pipeline as locomotion tasks. A domain-randomized motion tracking policy serves as the baseline. We evaluate both policies on three representative motions under impulse disturbances. The success rates are reported in Table~\ref{tab:ExpressiveMotion}.

\vspace{-0.15in}
\section{Limitation}
\label{sec:limitaion}
\vspace{-0.1in}
Our approach has three main limitations. First, policy transitions are governed by a learned discrete decision module, while the low-level controllers remain fixed. A jointly optimized hierarchical framework may enable smoother and more adaptive transitions. Second, the humanoid robot used in real-world deployment has only 19 degrees of freedom, which limits whole-body coordination and constrains the expression of complex recovery behaviors. Third, the recovery policy is trained in simulated dynamic scenarios that may not fully reflect the diversity and complexity of real-world disturbances. Incorporating more realistic perturbations, such as those generated by adversarial policies, could further improve recovery ability.
\vspace{-0.15in}
\section{Conclusion}
\label{sec:Conclusion}
\vspace{-0.1in}
We introduce HWC-Loco, a hierarchical control framework for humanoid robots that incorporates an embedded safety recovery mechanism. This framework has been validated across various tasks and scenarios, demonstrating exceptional effectiveness, robustness, naturalness, and scalability. Notably, the safety mechanism in HWC-Loco extends beyond locomotion, enabling reliable performance in complex tasks through a dynamic task-safety balance. This ensures robust operation in real-world deployments, positioning HWC-Loco as a foundational solution for safety-critical applications such as industrial automation and assistive robotics.
A promising direction for future research is integrating HWC-Loco with loco-manipulation skills, enabling safety-critical control across a broader range of tasks involving different objects.

\bibliography{main}
\bibliographystyle{unsrtnat}



\newpage
\appendix

\section{Addtional Details of HWC-Loco}
\label{Appendix:A}


\subsection{Motion Retargeting} 
\label{Appendix:Motion Retargeting}

Our motion data is derived from the CMU Mocap dataset \cite{cmudataset}, which exclusively encompasses human locomotion data, featuring a diverse array of walking, jogging, and running styles. In aggregate, we have loaded 318 motion sequences, totaling 3729.18 seconds in duration. Our motion retargeting methodology closely follows the approach outlined in \cite{Cheng2024Exbody}. We have meticulously aligned the humanoid robot's skeleton with that of humans, incorporating adjustments such as rotation, translation, and scaling. Additionally, we map human joints onto the humanoid robot's structure. Specifically, the human motion data is adapted to the humanoid robot's framework by mapping local information. Both the Unitree H1 and G1 robots possess shoulder and hip joints that are functionally equivalent to spherical joints. During the retargeting process, the three hip and shoulder joints are treated as a single spherical joint, corresponding to the human's spherical joint. For one-dimensional joints such as the elbow and torso, the rotation angle is projected onto the corresponding rotation axis of the joint angles.



\begin{table}[h]
\centering
\begin{minipage}{0.45\textwidth}
\centering
\caption{Double-DQN Parameters}
\begin{tabular}{l|l}
\toprule
Parameter & Value \\ 
\midrule
Batch Size & 128 \\ 
Learning Rate & 1e-4 \\ 
Gamma & 0.99 \\ 
Max Grad Norm & 1.0 \\ 
Replay Buffer Capacity & 2000000 \\ 
Switch Penalty Coefficient & 0.001 \\ 
Target Update Frequency & 50 \\ 
Optimizer & Adam \\ 
Loss Function & MSELoss \\ 
Initial Epsilon & 0.1 \\ 
Minimum Epsilon & 0.001 \\ 
Epsilon Decay Rate & 0.999 \\ 
\bottomrule
\end{tabular}
\end{minipage}
\hspace{0cm}  
\begin{minipage}{0.45\textwidth}
\centering
\caption{DQN Network Structure}
\begin{tabular}{l|l}
\toprule
Layer & Details \\ 
\midrule
Layer 1 & [input\_dim, 256] \\ 
Layer 2 & Activation: ReLU \\ 
Layer 3 & Regularization: Dropout \\ 
Layer 4 & [256, 128] \\ 
Layer 5 & [128, 64] \\
Layer 6 & Activation: ReLU \\ 
Layer 7 & [64, output\_dim] \\ 
\bottomrule
\end{tabular}
\end{minipage}
\end{table}



\subsection{Implementation Details}\label{Appendix:implementation}
\noindent\textbf{Proprioception:} For the Unitree H1, proprioception \(\boldsymbol{o}_{t}^p\in \mathbb{R}^{65}\), which denotes the internal state of the robot, including the base angular velocity \(\boldsymbol{w}_t \in \mathbb{R}^3\), base roll \(\boldsymbol{r}_t \in \mathbb{R}^1\), base pitch \(\boldsymbol{p}_t \in \mathbb{R}^1\), degrees of freedom (DOF) positions \(\boldsymbol{q}_t \in \mathbb{R}^{19}\), DOF velocities \(\dot{\boldsymbol{q}}_t \in \mathbb{R}^{19}\), previous actions \(\boldsymbol{a}_{t-1} \in \mathbb{R}^{19}\), and projected gravity \(\boldsymbol{g}_t \in \mathbb{R}^3\). The projected gravity refers to the component of gravity expressed in the robot's local coordinate system. For the Unitree G1, proprioception \(\boldsymbol{o}_{t}^p\in \mathbb{R}^{77}\), includes base angular velocity \(\boldsymbol{w}_t \in \mathbb{R}^3\), base roll \(\boldsymbol{r}_t \in \mathbb{R}^1\), base pitch \(\boldsymbol{p}_t \in \mathbb{R}^1\), DOF positions \(\boldsymbol{q}_t \in \mathbb{R}^{23}\), DOF velocities \(\dot{\boldsymbol{q}}_t \in \mathbb{R}^{23}\), previous actions \(\boldsymbol{a}_{t-1} \in \mathbb{R}^{23}\), and projected gravity \(\boldsymbol{g}_t \in \mathbb{R}^3\). As with the H1, the projected gravity denotes the component of gravity in the robot's local coordinate system.

\textbf{Action Space:} The policy outputs continuous actions $\mathbf{a}_t \in \mathbb{R}^n$, which are utilized as target positions for a PD controller to compute joint torques. The actions correspond to the robot's actuated degrees of freedom: specifically, $\mathbf{a}_t \in \mathbb{R}^{19}$ (19 DoF) for the Unitree H1 and $\mathbf{a}_t \in \mathbb{R}^{23}$ (23 DoF) for the Unitree G1.

\begin{figure}[htbp]
    \includegraphics[width=1.0\linewidth]{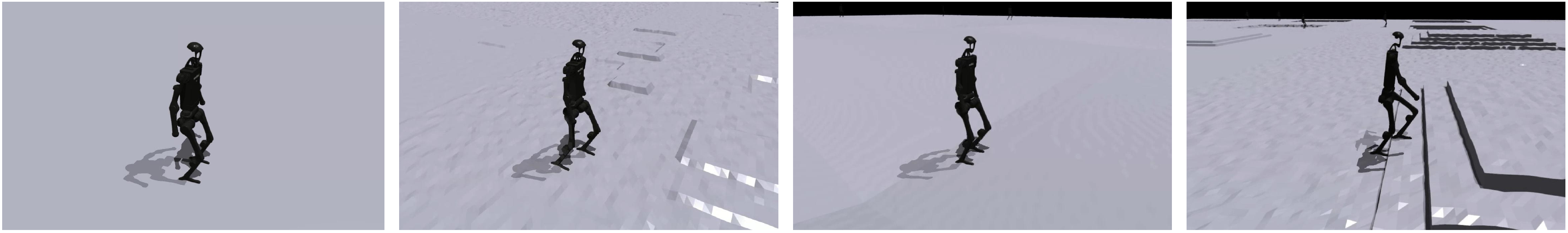}
    \caption{Visualization of different terrains. From left to right, the terrains are flats, obstacles, slopes, and stairs}
    \label{fig:Terrain}
\end{figure}

\textbf{Terrain details:} Terrains in the training environment simulated by Isaac Gym~\cite{makoviychuk2021isaac} include flats, obstacles, slopes, and stairs. An example of these terrains is shown in Figure~\ref{fig:Terrain}.

\noindent\textbf{Low-level Policy Training:} For all the low-level policies training, the commands are sampled from ranges: $v_{\text{x}}^c \sim [-0.6, 2.5]$, $v_{\text{y}}^c \sim [-0.6, 0.6]$ and $w_{\text{z}}^c \sim [-1.0, 1.0]$. 
During the initial training of the humanoid robot on uneven terrain, the robot often remains stationary due to the high difficulty of navigating complex surfaces. To address this, we introduce a terrain curriculum method \cite{makoviychuk2021isaac}. The training terrain consists of various types, including flat planes, rough surfaces, steps, and slopes. As the robot achieves a goal-tracking performance of 70\% of the commanded velocity, the terrain difficulty increases. Conversely, if the goal-tracking performance drops below 40\% of the commanded velocity, the terrain becomes easier.

\noindent\textbf{Learning for Policy Transition:} High-level planning policy is utilized to switch between low-level policies, ensuring compliance with robustness constraints. For the High-level policy, the input is the same set of observations as used by the low-level policies, with the output being a two-dimensional Q-value. During training, two trained low-level policies are loaded and rolled out to generate training data for optimizing the high-level policy. 
We train this high-level policy in the simulation environment mentioned detailed in Section~\ref{subsec:low-level-Recovery-policy}, focusing on locomotion goals as the primary task. The objective is to enable the robot to track goal commands across a variety of terrains. The details of the training process is in Algorithm \ref{alg:double_dqn_training}

\begin{algorithm}[H]
   \caption{Training for High-level Policy using Double-DQN}
   \label{alg:double_dqn_training}
\begin{algorithmic}
   \STATE {\bfseries Input:} Environment $env$, Selector network $S$, Target network $S_{\text{target}}$, Goal-tracking policy $L$, Recovery policy $R$
   \STATE Initialize: Learning rate $\alpha$, Discount factor $\gamma$, Exploration rate $\epsilon$, Replay buffer $B$, Target update frequency $f_{\text{target}}$, Switch penalty $\lambda_{\text{switch}}$, Exploration decay $\epsilon_{\text{decay}}$, Minimum exploration $\epsilon_{\text{min}}$
   \STATE $S_{\text{target}} \gets S$ \COMMENT{Initialize target network}

   \REPEAT
       \STATE Observe initial state $s$ from $env$
       \STATE Reset cumulative reward $R_{\text{total}}$, step count $n$, previous action $a_{\text{prev}} \gets \text{None}$

       \WHILE{not done}
           \STATE Compute $Q(s, \cdot) \gets S(s)$, select action $a$ with $\epsilon$-greedy
           \STATE Compute switch penalty: $r_{\text{penalty}} \gets \lambda_{\text{switch}} \cdot \mathbf{1}[a \neq a_{\text{prev}}]$
           \STATE Execute $a$, observe reward $r$, next state $s'$, and done signal $d$
           \STATE Store $(s, a, r + r_{\text{penalty}}, s', d)$ in $B$
           \STATE Update $s \gets s'$, $a_{\text{prev}} \gets a$
           \STATE $R_{\text{total}} \gets R_{\text{total}} + r$, $n \gets n + 1$

           \IF{$B$.size() $\geq$ batch size}
               \STATE Sample $(s, a, r, s', d)$ from $B$
               \STATE $a' \gets \arg\max_{a'} S(s', a')$ \COMMENT{Online network selects action}
               \STATE $Q^{\text{target}}(s', a') \gets S_{\text{target}}(s', a')$ \COMMENT{Target network evaluates}
               \STATE Compute target: $Q^{\text{target}}(s, a) \gets r + \gamma (1 - d) Q^{\text{target}}(s', a')$
               \STATE Update $S$ by minimizing:  
               $\mathcal{L} = \left( S(s, a) - Q^{\text{target}}(s, a) \right)^2$
           \ENDIF

           \IF{$n \bmod f_{\text{target}} = 0$}
               \STATE $S_{\text{target}} \gets S$ \COMMENT{Sync target network}
           \ENDIF
       \ENDWHILE

       \STATE Decay exploration rate: $\epsilon \gets \max(\epsilon \cdot \epsilon_{\text{decay}}, \epsilon_{\text{min}})$
   \UNTIL{convergence criteria met}
   \STATE {\bfseries Output:} Trained Selector Network $S$
\end{algorithmic}
\end{algorithm}

\noindent\textbf{Reward Details:} To better facilitate efficient humanoid locomotion, our task reward is defined as~\cite{gu2024humanoid,gu2024advancing}:
\begin{align*}
r_\mathfrak{T} &= \alpha_{1} \exp\left(-\frac{\|v_{\text{xy}}^c - v_{\text{xy}}\|_2^2}{\sigma_{\text{lin}}^2}\right) 
+ \alpha_{2}\exp\left(-\frac{\|w_{\text{z}}^c - w_{\text{z}}\|_2^2}{\sigma_{\text{ang}}^2}\right)
\end{align*}
where 1) \(v_{\text{xy}}^c\) denotes the commanded linear velocity along the \(x\) and \(y\) axes, while \(w_{\text{z}}^c\) is the commanded yaw velocity. 2) \(v_{\text{xy}}\) and \(w_{\text{z}}\) represent the corresponding base velocities of the humanoid robot. 3) \(\alpha_{1}\) and \(\alpha_{2}\) indicate the hyperparameters that are utilized to adjust the importance of the different velocity terms. 4) \(\sigma_{\text{lin}}^2\) and \(\sigma_{\text{ang}}^2\) control the precision of the expected command tracking. Smaller values of  \(\sigma_{\text{lin}}^2\) and \(\sigma_{\text{ang}}^2\) 
enhance the precision of command tracking for humanoid robots. However, they may hinder reward acquisition during the initial stages of training.

The recovery policy shares the same task reward formulation as the goal-tracking policy. However, it uses larger values of the \(\sigma_{\text{lin}}^2\) and \(\sigma_{\text{ang}}^2\), which implies a greater tolerance for velocity tracking errors. As a result, this reward term serves as a back-tracking reward for the safety recovery mechanism, encouraging it to return to a stable goal-tracking state. To further promote stable posture restoration and enable smooth transitions back to the goal-tracking policy, we introduce an additional stand reward, defined as:
\[
r_{\text{stand}} = \|\mathbf{q}_t - \mathbf{q}_{\text{default}}\|_2^2.
\]
where $\mathbf{q}_{\text{default}}$ represents a default standing pose.

For safety realistic deployment, we add some energy and safety reward for the low-level policies training. These rewards are shown in Table. 

\begin{table}[htbp]
    \centering
    \caption{Safety \& Energy Reward}
    \begin{tabular}{l|c|c}
        \toprule
        Term & Expression & Weight \\
        \midrule
        DoF position limits & $\mathbf{1}(\mathbf{q}_t \notin [\mathbf{q}_{\min}, \mathbf{q}_{\max}])$ & $-1e^{2}$ \\
        DoF acceleration & $\|\ddot{\mathbf{q}}_t\|_2^2$ & $-1e^{-7}$ \\
        DoF velocity & $\|\dot{\mathbf{q}}_t\|_2^2$ & $-5e^{-4}$ \\
        Action rate & $\|\mathbf{a}_t - \mathbf{a}_{t - 1}\|_2^2$ & $-2e^{-3}$ \\
        Torque & $\|\tau_t\|$ & $-1e^{-5}$ \\
        Collision & $\sum_{i\in \text{contact}} \mathbf{1}\{\|F_i\| > 1.0\}$ & $-10$ \\
        \bottomrule
    \end{tabular}
\end{table}

\subsection{PPO} 
\label{Appendix:PPO}
In order to enhance the understanding of historical observation information, we have made some improvements to the Actor-network. We first process this information using an encoder network to preliminarily extract the features of the observation values at each time step, and then use a merger network to integrate the observation values at different time steps. Finally, we obtain fixed-dimensional feature information, which is then input into the Actor backbone network together with other information.

\begin{table}[H]
\centering
\caption{PPO hyperparameters}
\begin{tabular}{l|l}
\toprule
Hyperparameter & Value \\ 
\midrule
Actor Layer & [512, 256, 128] \\ 
Critic Layer & [512, 256, 128] \\
Activation Function & ELU \\
Discount Factor & 0.99 \\ 
GAE Parameter & 0.95 \\ 
Epochs per Rollout & 5 \\ 
Minibatch & 4 \\ 
Entropy Coefficient  & 0.005 \\ 
Value Loss Coefficient ($\alpha_{1}$) & 1.0 \\ 
Clip Range & 0.2 \\ 
Learning Rate & 2.e-4 \\ 
Environment Steps per Rollout & 24 \\ 
\bottomrule
\end{tabular}
\end{table}

\subsection{VAE}
\label{Appendix:VAE}
\begin{figure}[H]
    \centering
    \begin{minipage}{0.48\linewidth}
        \centering
        \includegraphics[width=\linewidth]{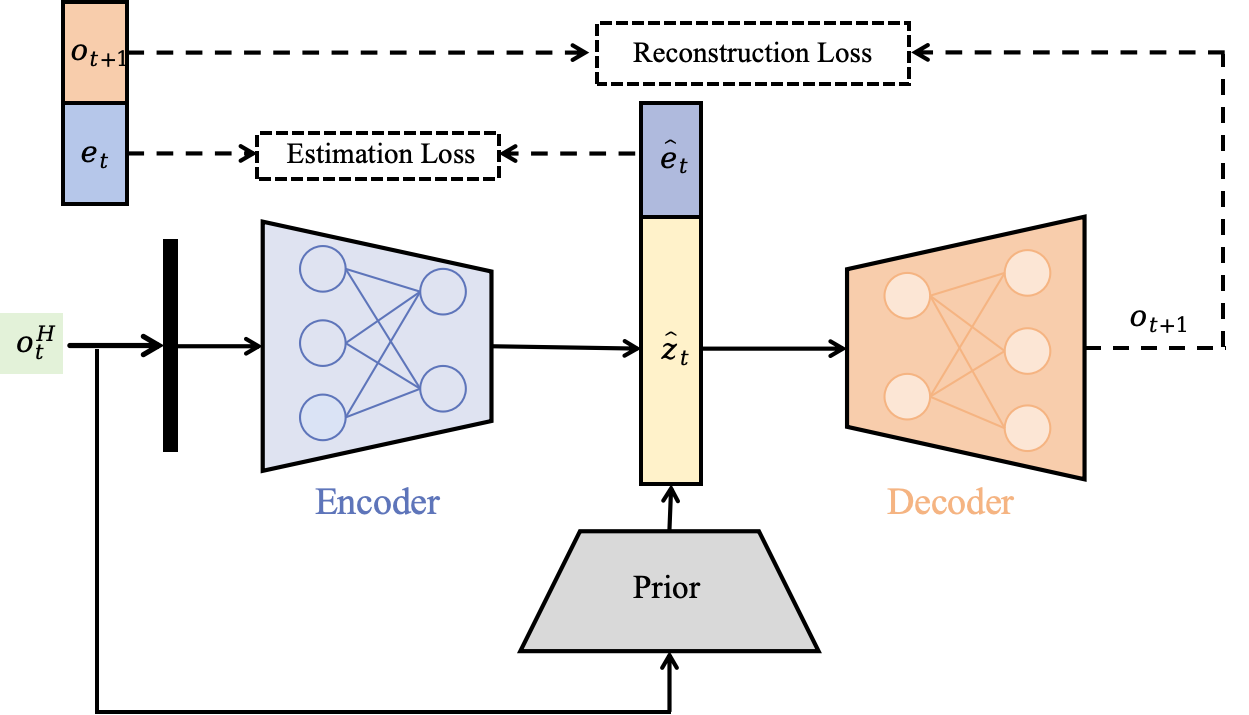}
        \caption{VAE Structure}
        \label{fig:VAE}
    \end{minipage}
    \hfill
    \begin{minipage}{0.48\linewidth}
        \centering
        \captionof{table}{VAE hyperparameters}
        \begin{tabular}{l|l}
            \toprule
            Hyperparameter & Value \\ 
            \midrule
            Encoder Layer & [256, 128] \\ 
            Decoder Layer & [256, 128, 64] \\
            Activation Function & ELU \\
            KL Coefficient  & 1.0 \\ 
            Learning Rate & 1.e-4 \\ 
            \bottomrule
        \end{tabular}
    \end{minipage}
\end{figure}

The VAE Network consists of an encoder and a decoder. The encoder takes historical observations as input and outputs latent representation and estimation of priviledged information. The decoder takes latent as input and outputs the reconstruction of the next observation. The structure of VAE is shown in Figure~\ref{fig:VAE}

\subsection{Discriminator}
\label{Appendix:Discriminator}

The input to the discriminator \( s^d \) is defined as \([q, v, w, r, p, k]\), where \( k \) represents the local positions of key points. Specifically, we define 12 key points, including the hips, shoulders, hands, ankles, knees, and elbows. For the Unitree H1, the total input dimension is 63, while for the Unitree G1, it is 67.

The discriminator's hyperparameters are shown in Table. \ref{tab:Discriminator}.

\begin{table}[h]
\centering
\caption{Discriminator hyperparameters}
\label{tab:Discriminator}
\begin{tabular}{l|l}
\toprule
Hyperparameter & Value \\ 
\midrule
Discriminator Hidden Layer Dim & [1024,512] \\ 
Replay Buffer Size & 500000 \\ 
Expert Buffer Size & 200000 \\ 
Expert Fetch Size & 512 \\ 
Learning Batch Size & 4096 \\ 
Learning Rate & 2e-5 \\ 
Gradient Penalty Coefficient & 1.0 \\ 
lamda & 1.0 \\ 
\bottomrule 
\end{tabular}
\end{table}


\subsection{Saliency Analysis of Observations}
\label{Appendix:Saliency Analysis}
\begin{figure}[htbp]
    \centering
    \includegraphics[width=0.48\linewidth]{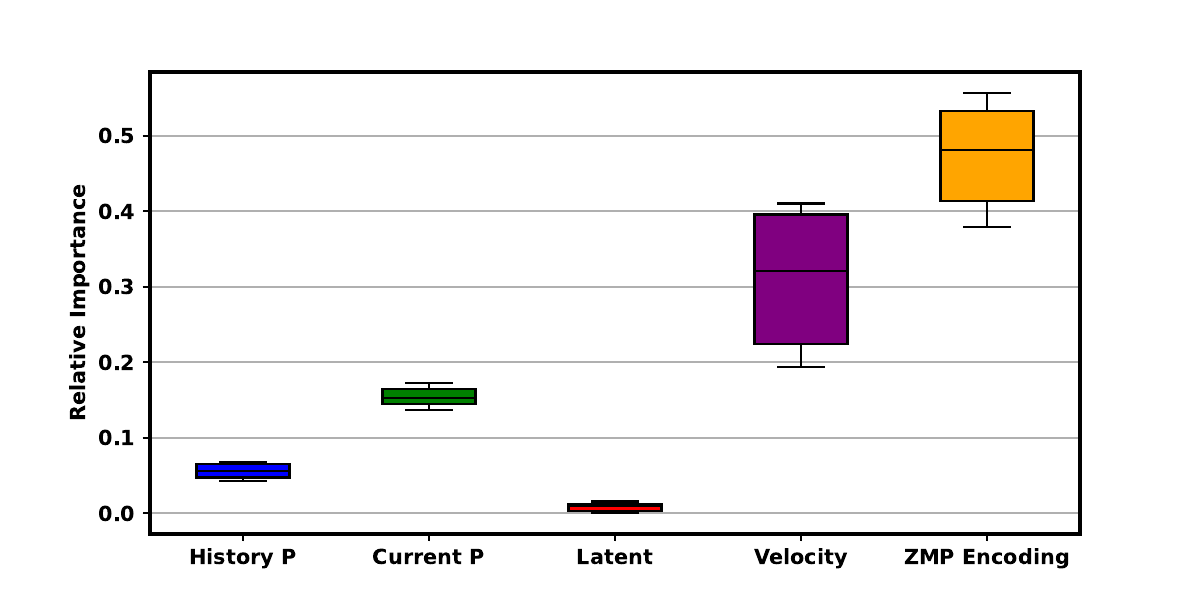}
    \includegraphics[width=0.48\linewidth]{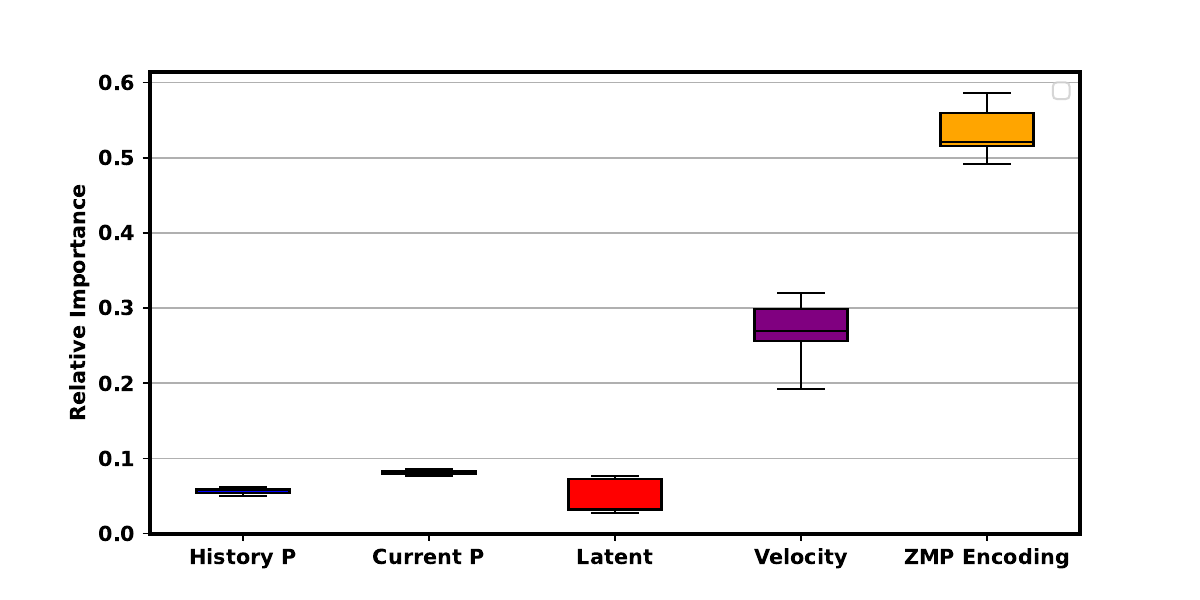}
    \caption{Importance of Observations}
    \label{fig:Saliency Analysis}
\end{figure}

We adopt an integrated gradients method similar to \cite{wang2024toward} to assess the importance of different parts of the observation to the policy, as shown in Fig.\ref{fig:Saliency Analysis}. 

For the recovery policy, the Zero Moment Point (ZMP) encoding proves to be the most critical factor, followed by the estimated velocity. The robot’s current and historical proprioceptive states also contribute substantially to the policy’s decision-making process. In contrast, the environmental latent representation exhibits only a marginal effect. This hierarchy of importance reflects the necessity of strictly adhering to ZMP constraints to maintain dynamic stability and ensure the safety of the robot during recovery.

For the high-level policy, the Zero Moment Point (ZMP) encoding remains the most influential component, indicating that it provides critical information for decision-making and substantially enhances the policy’s awareness of the robot’s stability. This observation implies that the high-level policy prioritizes stability-relevant features over task-irrelevant ones, which is advantageous for promoting generalization across diverse scenarios. Moreover, the environmental latent variable exhibits greater importance compared to that in the recovery policy, suggesting that high-level decisions are more sensitive to environmental factors such as terrain variations.

\subsection{Extreme Initialization}
\label{Appendix:Extreme Initialization}
To simulate extreme scenarios that may occur in real-world deployment, we identify three primary causes of such cases: external disturbances, hardware malfunctions, and malicious commands. Additionally, we retain domain randomization to enhance the generalization of physical dynamics within the simulation environment. The specific details are outlined as follows:

To enhance robustness and generalization in simulation, we simulate both extreme deployment scenarios and apply domain randomization. The extreme scenarios include external disturbances, sensor and actuator failures, and adversarial commands, while domain randomization targets physical variability in the environment.

1) \textbf{External disturbances:} We apply external disturbances to every component of the robot. The external forces and torques are applied along the x, y, and z axes, with each direction sampled from the range of -200 N to 200 N for forces and -200 Nm to 200 Nm for torques. These disturbance forces and torques are updated at a frequency of 1 Hz.

2) \textbf{Hardward magnification :} Random, high-intensity noise is introduced to the proprioceptive information and PD gains. 
The noise added to the proprioceptive data simulates sensor errors that may occur during real-world deployment. The noise applied to the PD gains simulates fluctuations in motor strength, reflecting potential issues in the robot's actuators. The parameters for this noise are detailed in Table \ref{tab:obs_noise}.

3) \textbf{Malicious velocity:} Velocity commands are resampled from the command range, with updates occurring at a frequency of 5 Hz. Notably, this resampling mechanism is activated once the terrain curriculum exceeds half of its total progression.

4) \textbf{Domain randomization  (for environment variability):} Domain randomization is applied to all the training environments. This technique introduces variability in the environment’s parameters, such as friction and gravity. The specifics of the domain randomization are provided in Table~\ref{tab:domain_rand}.

To improve the generalization of recovery ability, we utilize random initialization with extreme case initialization. The parameters are shown in Table~\ref{tab:init_params}

\begin{table}[h]
\centering
\caption{Intense Noise Parameters}
\label{tab:obs_noise}
\begin{tabular}{l|l|l}
\toprule
Parameter & Range & Operator \\ 
\midrule
DOF Position Noise & 0.04 & additive \\ 
DOF Velocity Noise & 0.30 & additive \\ 
Angular Velocity Noise & 1.00 & additive \\ 
IMU Noise & 0.40 & additive \\ 
Gravity Noise & 0.10 & additive \\ 
PD Gains & 0.20 & additive \\ 
\bottomrule
\end{tabular}
\end{table}

\begin{table}[h]
\centering
\begin{minipage}{0.45\textwidth}
\centering
\caption{Domain Randomization Parameters}
\label{tab:domain_rand}
\begin{tabular}{l|l|l}
\toprule
Parameter & Range & Operator \\ 
\midrule
Friction & [0.6, 2.0] & additive \\ 
Base Mass & [-1.0, 5.0] & additive \\ 
Base CoM & [-0.07, 0.07] & additive \\ 
Motor Strength & [0.8, 1.2] & scaling \\ 
Action Delay & [0, 5] & - \\ 
Link Mass & [0.7, 1.3] & scaling \\ 
Gravity & [-0.1, 0.1] & additive \\ 
\bottomrule
\end{tabular}
\end{minipage}
\hspace{0cm}  
\begin{minipage}{0.45\textwidth}
\caption{Random Initialization Parameters}
\label{tab:init_params}
\begin{tabular}{l|l}
\toprule
Parameter           & Range \\ 
\midrule
Linear Velocity (x)          & \([-1.0, 2.5]\)           \\
Linear Velocity (y)          & \([-1.0, 1.0]\)           \\
Linear Velocity (z)          & \([-0.4, 0.4]\)           \\
Angular Velocity             & \([-1.0, 1.0]\)           \\
DOF Position                 & \([0.2 \times \text{default}, 1.8 \times \text{default}]\) \\
DOF Velocity          & \([-0.2, 0.2]\)           \\
Pitch Angle                  & \([-0.25, 0.25]\)         \\
Yaw Angle                    & \([-0.4, 0.4]\)           \\
Position                     & \([-2, 2]\)              \\ 
\bottomrule
\end{tabular}
\end{minipage}
\end{table}

\begin{figure}[htbp]
    \centering
    \includegraphics[width=0.24\linewidth]{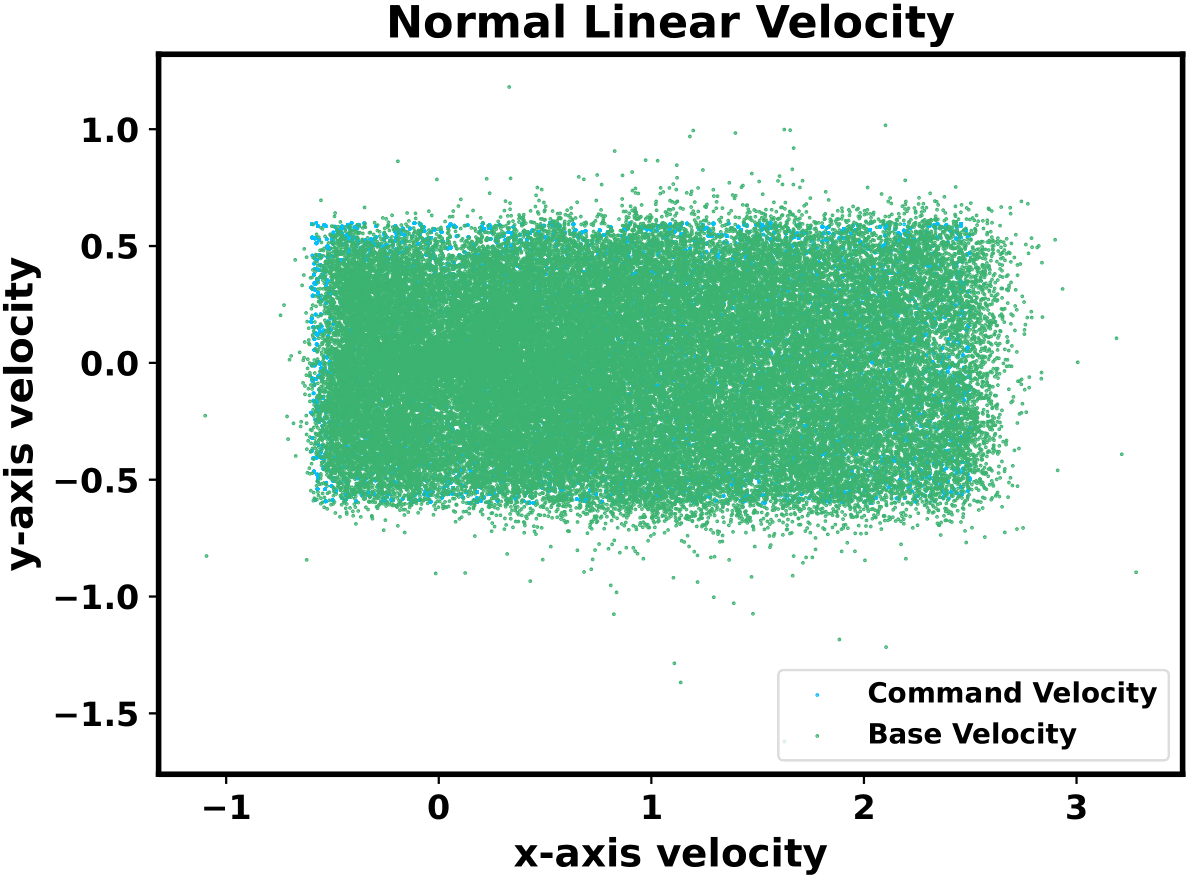}
    \includegraphics[width=0.24\linewidth]{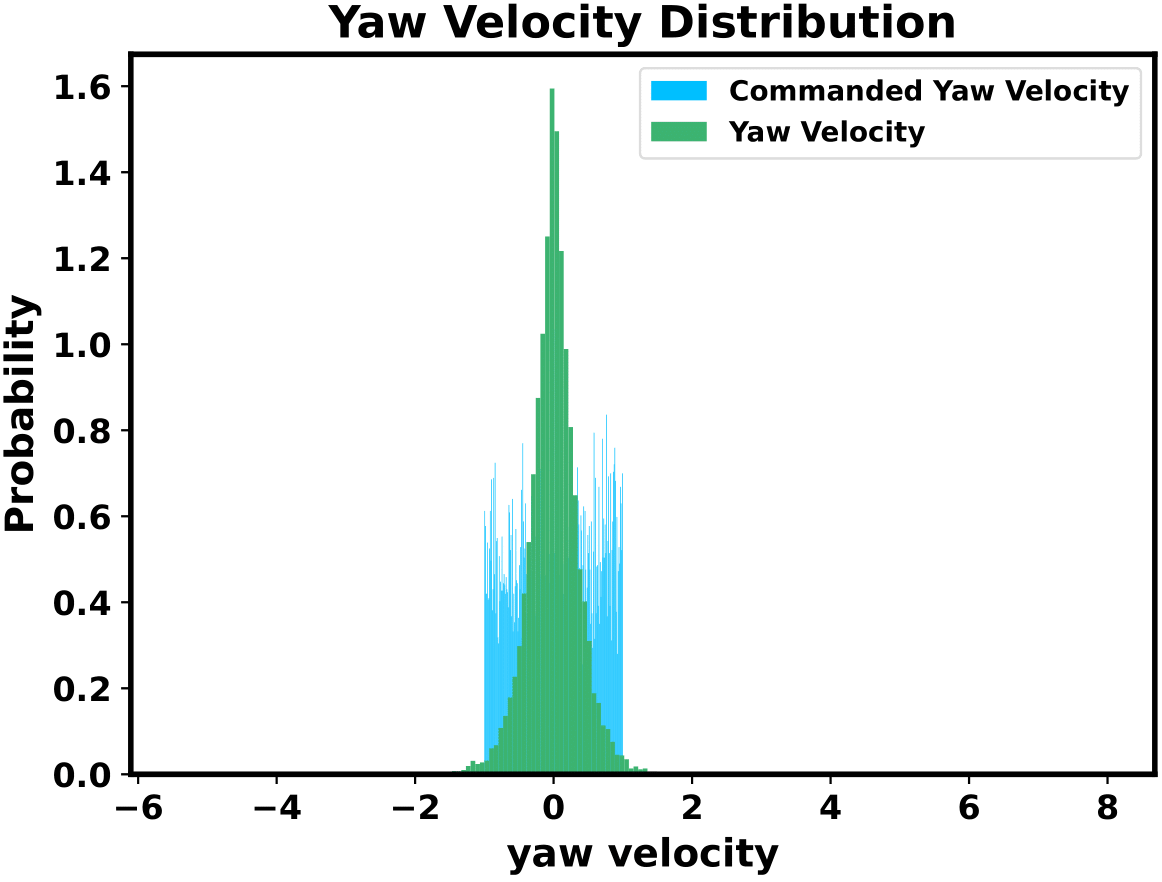}
    \includegraphics[width=0.24\linewidth]{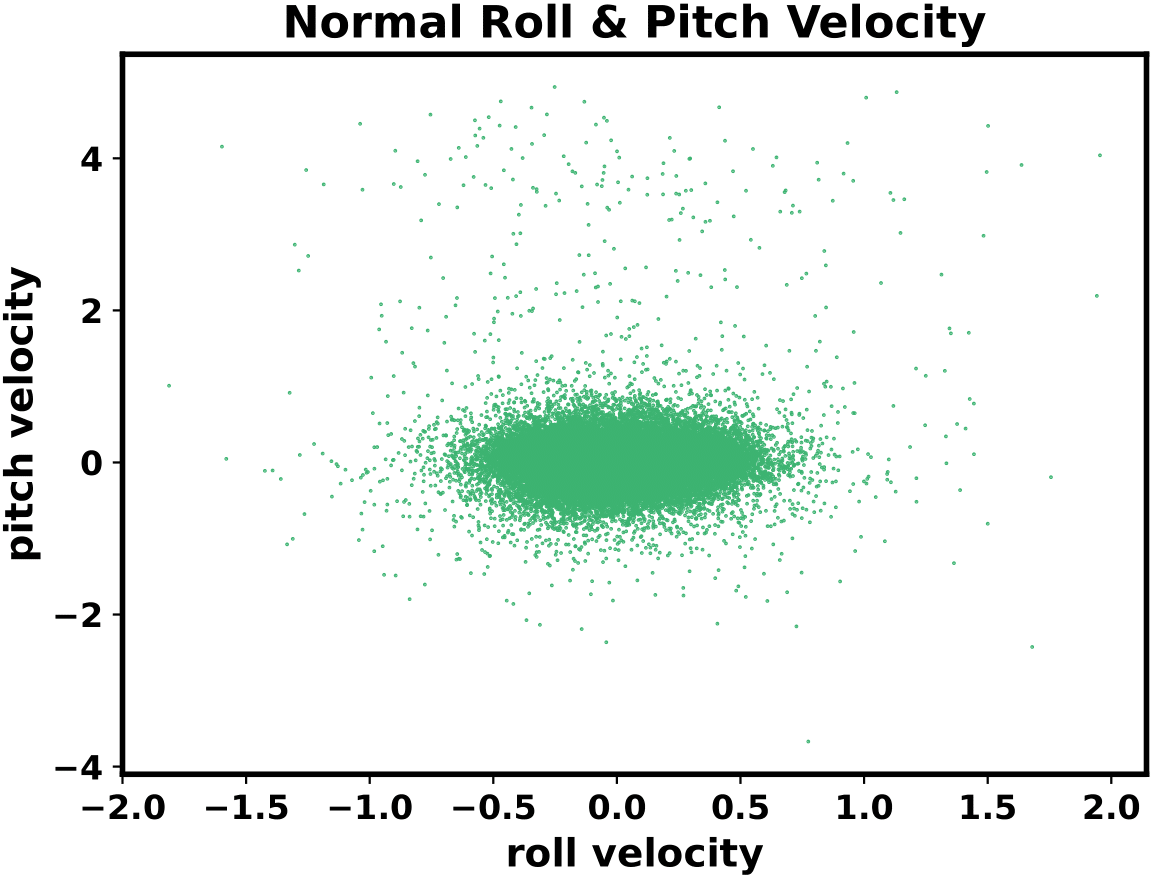}
    \includegraphics[width=0.24\linewidth]{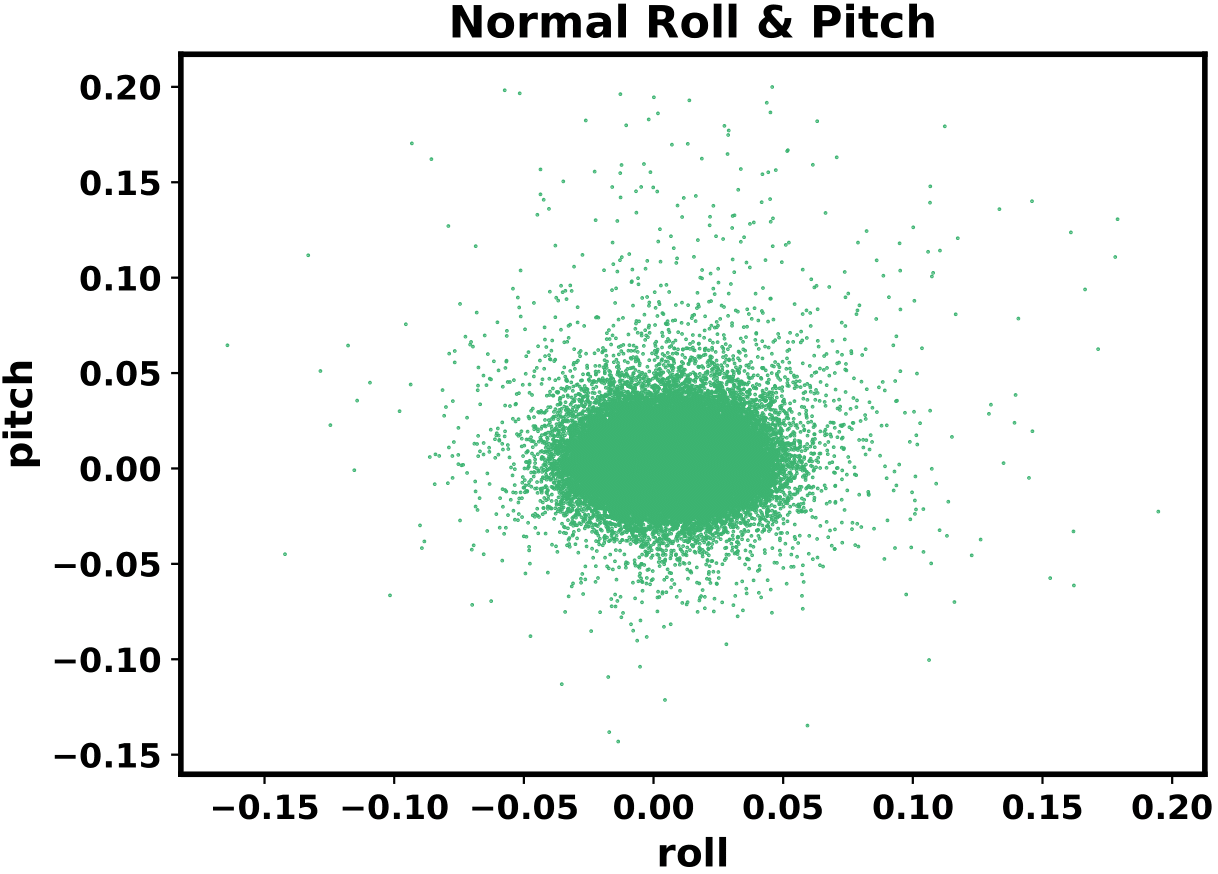}
    \caption{Normal State: Policy rollout’s state distribution in the original learning environment}
    \label{fig:Normal State}
\end{figure}

\begin{figure}[htbp]
    \centering
    \includegraphics[width=0.24\linewidth]{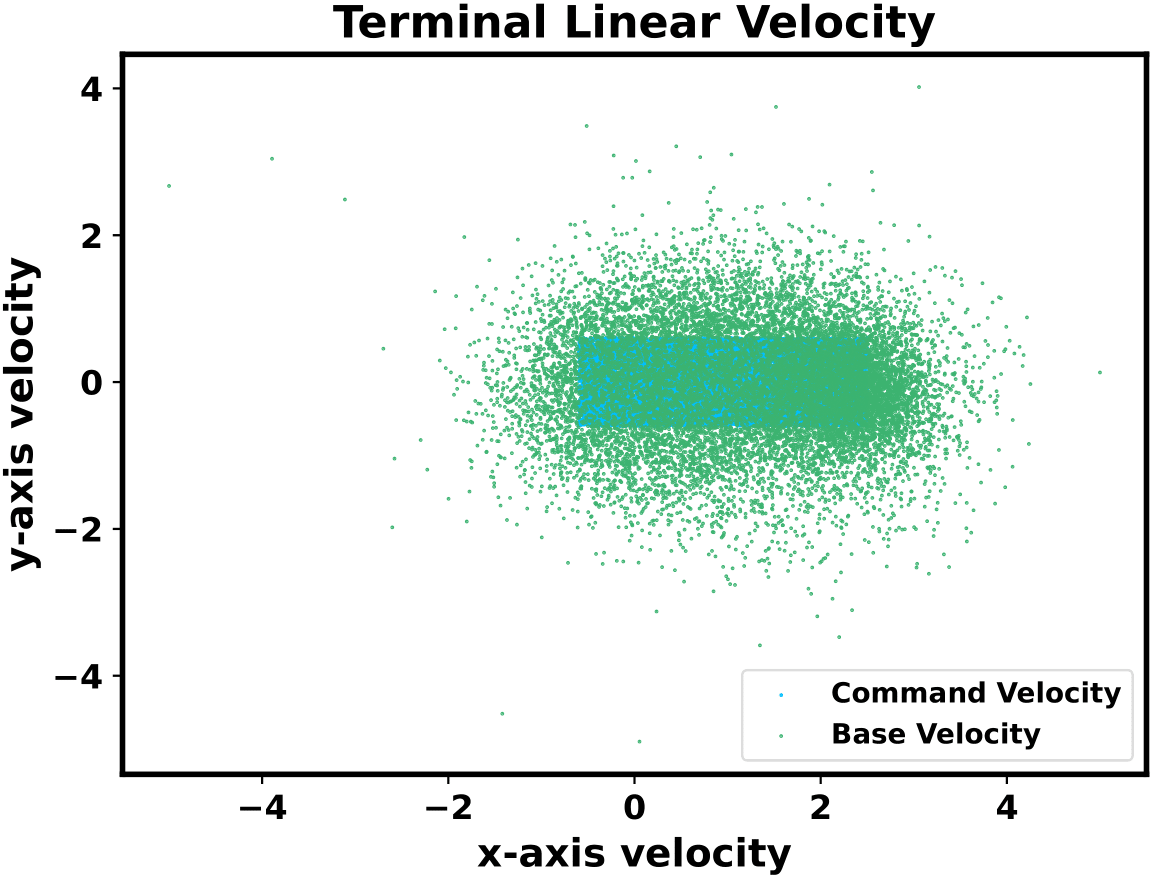}
    \includegraphics[width=0.24\linewidth]{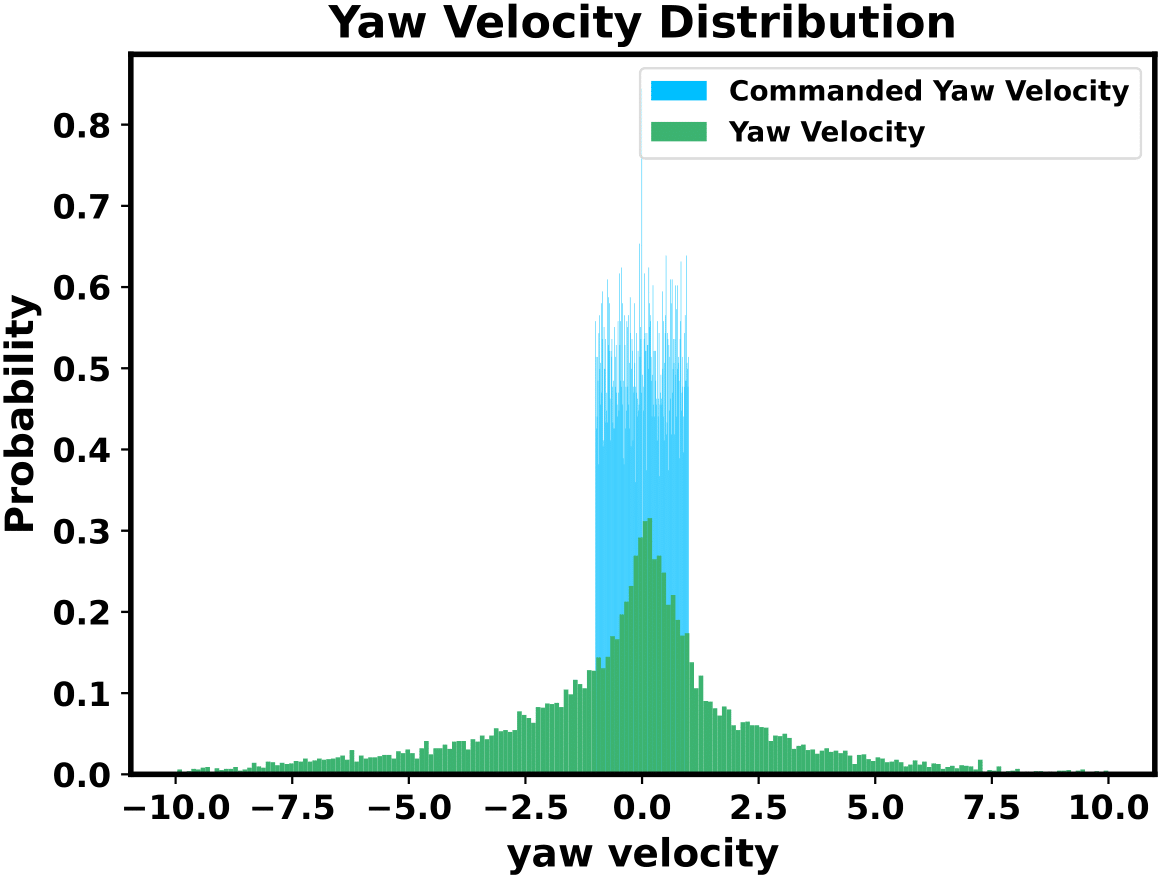}
    \includegraphics[width=0.24\linewidth]{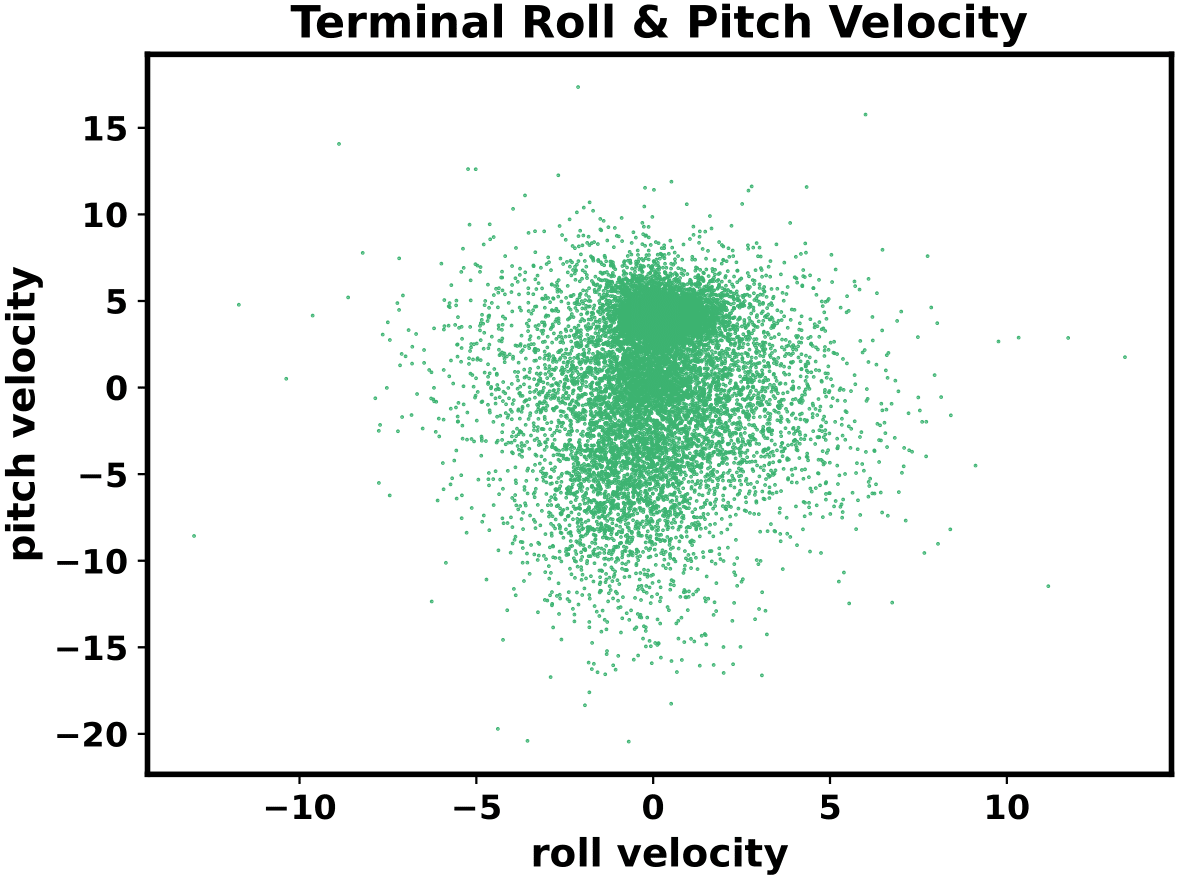}
    \includegraphics[width=0.24\linewidth]{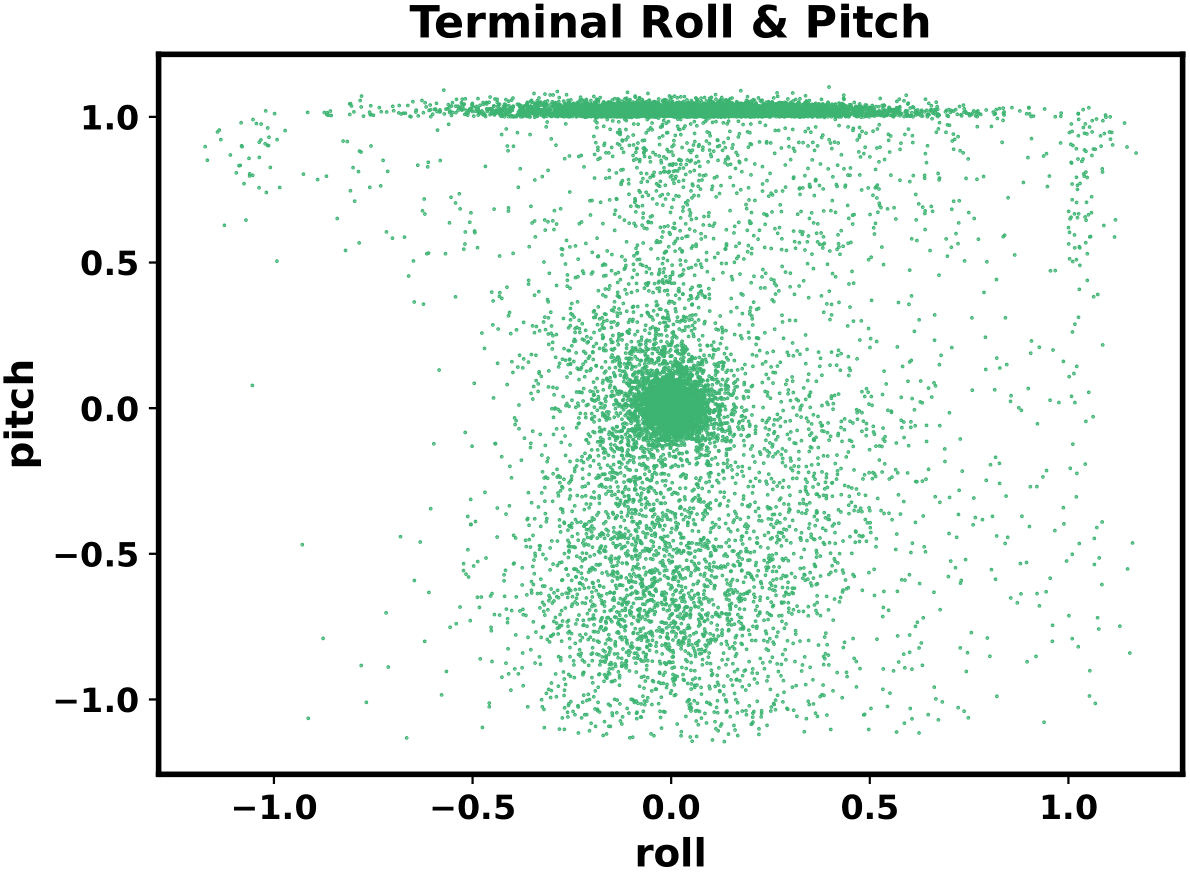}
    \caption{Extreme State: Policy’s state distribution in the extreme cases}
    \label{fig:Extreme State}
\end{figure}

\textbf{Extreme-Case Analysis:} We record nearly 60000 normal state data with randomly sampled commands in diverse terrain settings in \ref{fig:Normal State}.
We record nearly 10000 extreme cases in the $\trainsitionSet_{\alpha}^{L}$. Some of the terminal states are visualized in Fig. \ref{fig:Extreme State}. For the extreme states, the distribution of velocity is close to the command ranges. Roll pitch and their velocity are in a common range, which is near zeros. For the normal states, the distribution of linear velocity overlaps with the command velocity range, with a dense cluster around the high-velocity value of 2 m/s. Regarding yaw velocity, most values fall within the commanded range. However, there are instances of high speed, with some even approaching 10 m/s. This may be due to the robot’s extreme behaviors when encountering unseen situations influenced by external forces and random noise. Regarding the roll-pitch velocity, the pitch velocity shows a higher degree of fluctuation. Similarly, in the roll pitch distribution, for pitch, there is a concentrated distribution at a pitch of 1 m/s, which is one of the terminal conditions. These two figures suggest that the robot is tripped by obstacles during its motion and thus terminates, for instance, when it hits steps or gravel during high-speed motion.

\subsection{History Length Experiments}
\label{Appendix:History_Length}

We investigate the impact of observation history length on HWC-Loco’s performance. Setting $H = 10$ achieves the best overall trade-off across the three metrics. While increasing $H$ beyond 10 slightly improves the success rate, it leads to a noticeable decline in goal-tracking accuracy and human-likeness. This suggests that excessively long history may hinder the effective use of privileged information, resulting in more conservative and less adaptive behaviors.

\begin{table}[H]
\centering
\caption{Performance comparison with different history lengths ($H$)}
\label{tab:history_length}
\begin{tabular}{lccc}
\toprule
\textbf{History Length} & \textbf{Success Rate $\uparrow$} & \textbf{Goal-tracking $\uparrow$} & \textbf{Human-like $\downarrow$} \\
\midrule
1   & $95.13 \pm 0.51$ & $1.07 \pm 0.00$ & $3.26 \pm 0.06$ \\
10  & $96.73 \pm 0.40$ & $\mathbf{1.10 \pm 0.00}$ & $\mathbf{3.18 \pm 0.01}$ \\
20  & $96.53 \pm 0.40$ & $1.08 \pm 0.00$ & $3.18 \pm 0.02$ \\
40  & $\mathbf{97.38 \pm 0.33}$ & $1.02 \pm 0.00$ & $3.21 \pm 0.02$ \\
\bottomrule
\end{tabular}
\end{table}

\subsection{Training Cost} 
All policies are trained on an NVIDIA RTX 4090 GPU with 4,096 parallel environments. Our framework involves training three policies. The goal-tracking policy converges within approximately 10000 iterations (nearly 10 hours), while the recovery policy requires around 8000 iterations to converge (nearly 8 hours). The high-level policy demonstrates faster convergence, requiring approximately 6000 iterations (under 6 hours).

\subsection{Deployment Details} 
\label{Appendix:Deployment}
We utilize the Unitree H1 humanoid robot as our deployment platform. This full-sized humanoid robot weighs 47 kg, stands approximately 1.8 meters tall, and features 19 degrees of freedom. The control frequency is set to 100 Hz for both simulation and real-world deployment. The PD gains, characterized by stiffness and damping values, used in the real-world deployment are detailed in Table \ref{tab:PDgains}.

During deployment, inference operates through a hierarchical structure involving both high-level and low-level policies. To meet the strict latency and resource constraints of onboard systems, the policies are optimized for lightweight designs and exported JIT format. The high-level policy model size is below 2MB, and the low-level policies remain under 3MB. Those designs ensures that real-time inference at 100 Hz is feasible on the onboard computer of the Unitree H1 robot.

\begin{table}[h]
\centering
\caption{Stiffness, Damping and Torque Limit}
\label{tab:PDgains}
\begin{tabular}{l|l|l|l}
\toprule
Joint Names & Stiffness [N*m/rad] & Damping [N*m*s/rad] & Torque Limit (Nm) \\ 
\midrule
hip yaw & 200 & 5 & 170 \\ 
hip roll & 200 & 5 & 170 \\ 
hip pitch & 200 & 5 & 170 \\ 
knee & 300 & 6 & 255 \\ 
ankle & 40 & 2 & 34 \\ 
torso & 200 & 5 & 170 \\ 
shoulder & 30 & 1 & 34 \\ 
elbow & 30 & 1 & 18 \\ 
\bottomrule
\end{tabular}
\end{table}

\section{Experiment}
\label{Appendix:Experiment}




\subsection{H1 Experiments}
\label{Appendix:Deployment}

\begin{figure*}[h]
    \centering
    \includegraphics[width=0.95\linewidth]{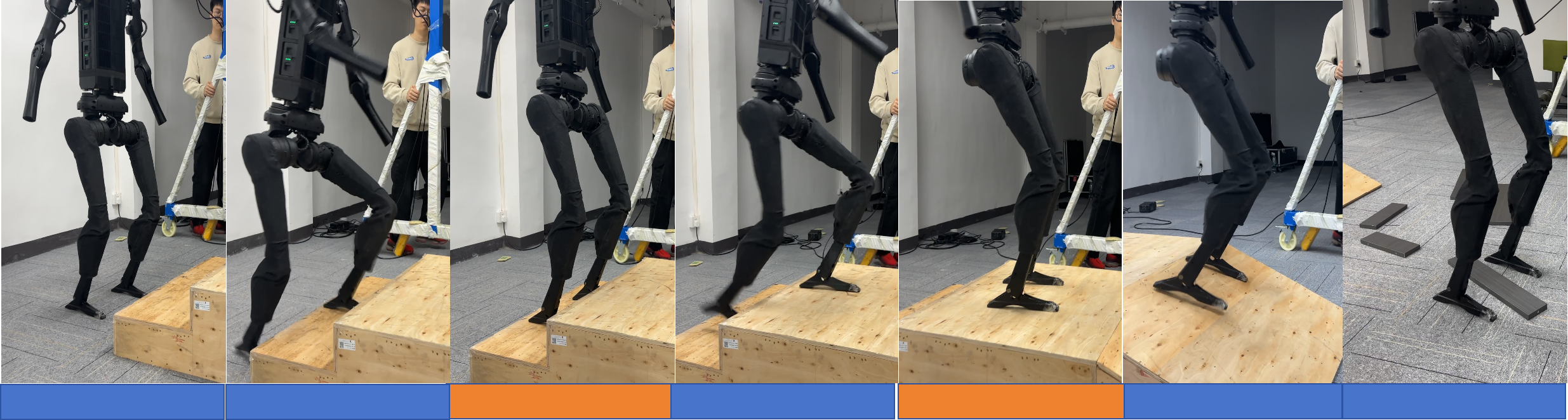}
    \caption{Climb Stairs Test. The blue segments indicate the activation of the goal-tracking policy, while the orange segments correspond to the safety recovery policy.}
    \label{fig:Climb_stairs}
\end{figure*}

\begin{figure*}[htbp]
    \centering

    \begin{minipage}[t]{0.95\linewidth}
        \centering
        \includegraphics[width=\linewidth]{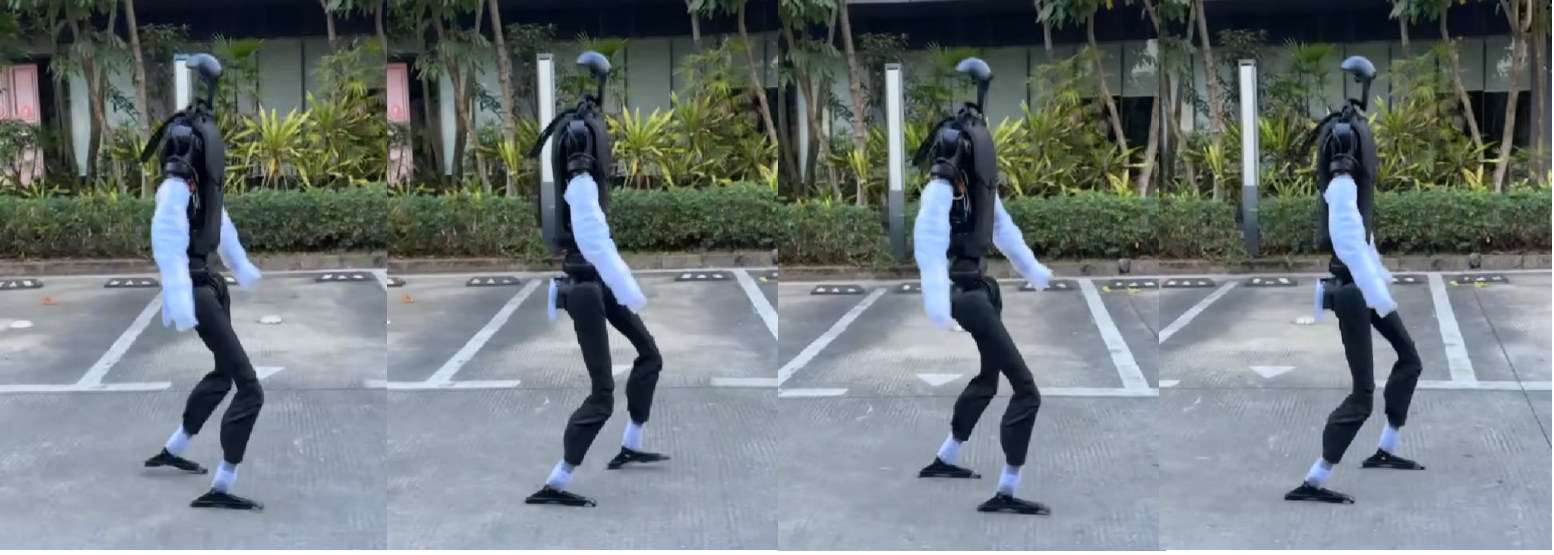}
        \caption*{(a) Walking on flat terrain.}
    \end{minipage}

    \begin{minipage}[t]{0.95\linewidth}
        \centering
        \includegraphics[width=\linewidth]{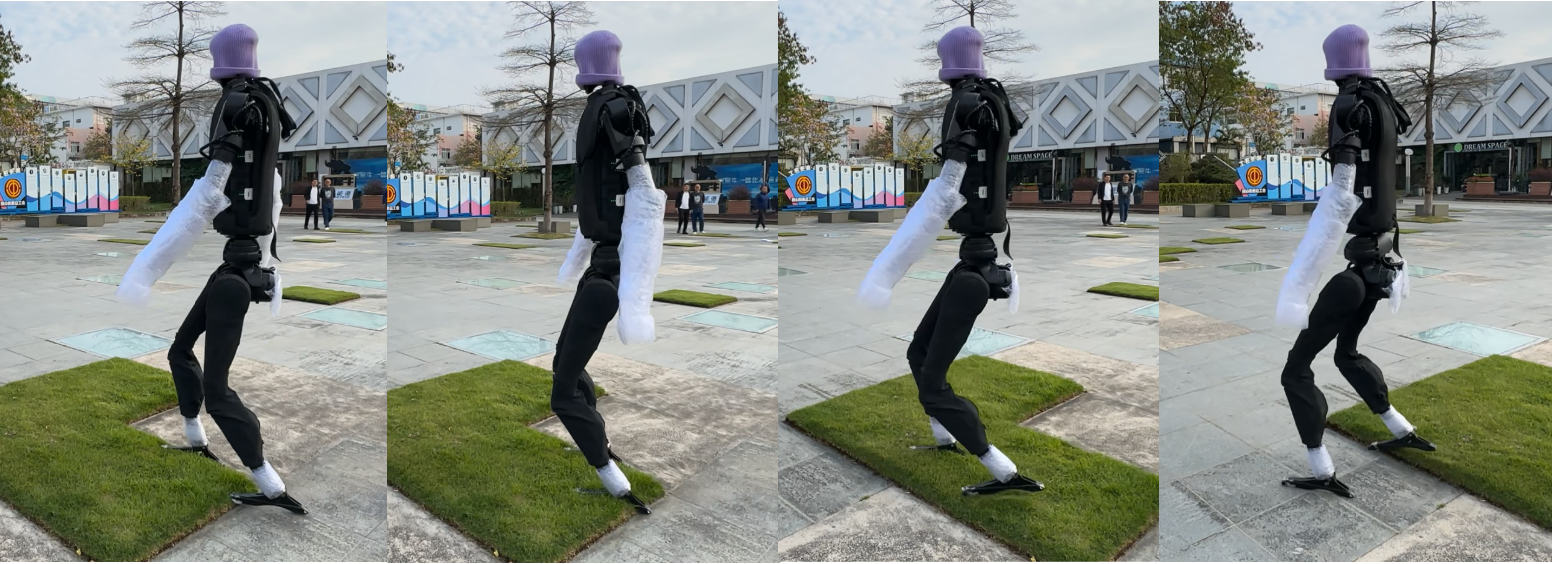}
        \caption*{(b) Traversing grassy surfaces.}
    \end{minipage}

    \begin{minipage}[t]{0.95\linewidth}
        \centering
        \includegraphics[width=\linewidth]{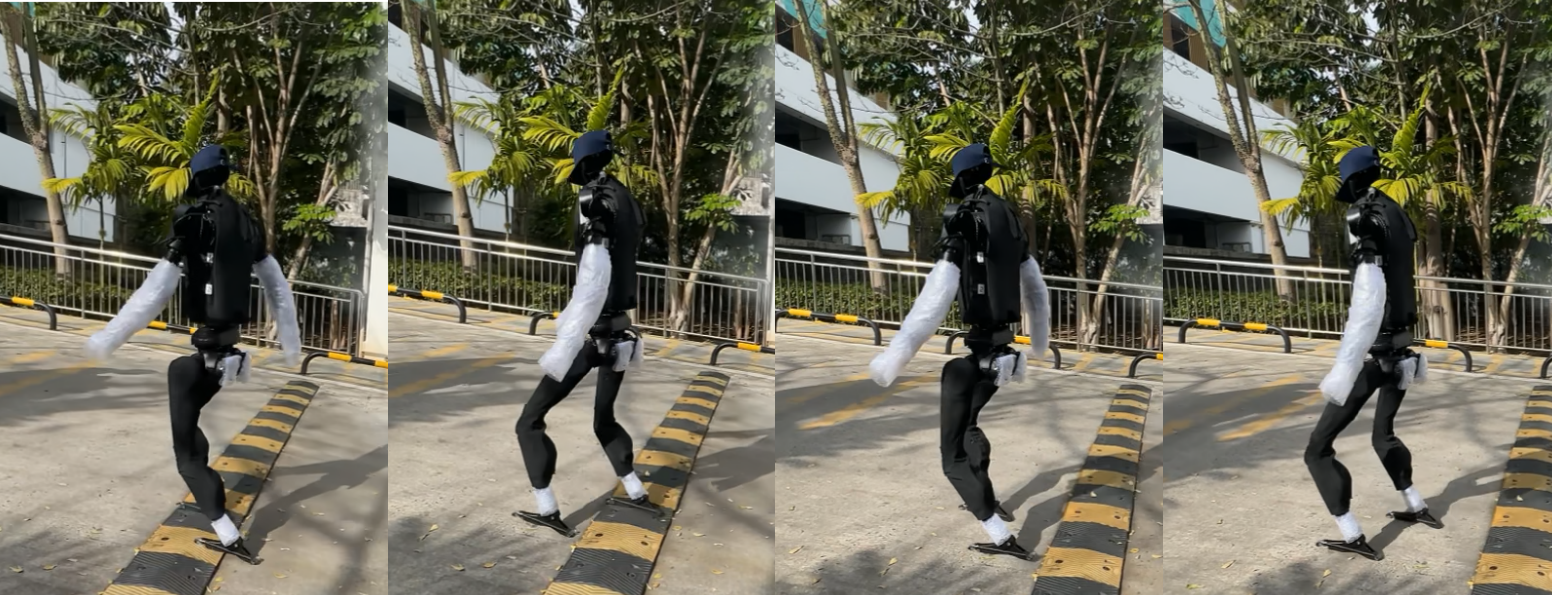}
        \caption*{(c) Climbing a slope.}
    \end{minipage}
    
    \begin{minipage}[t]{0.95\linewidth}
        \centering
        \includegraphics[width=\linewidth]{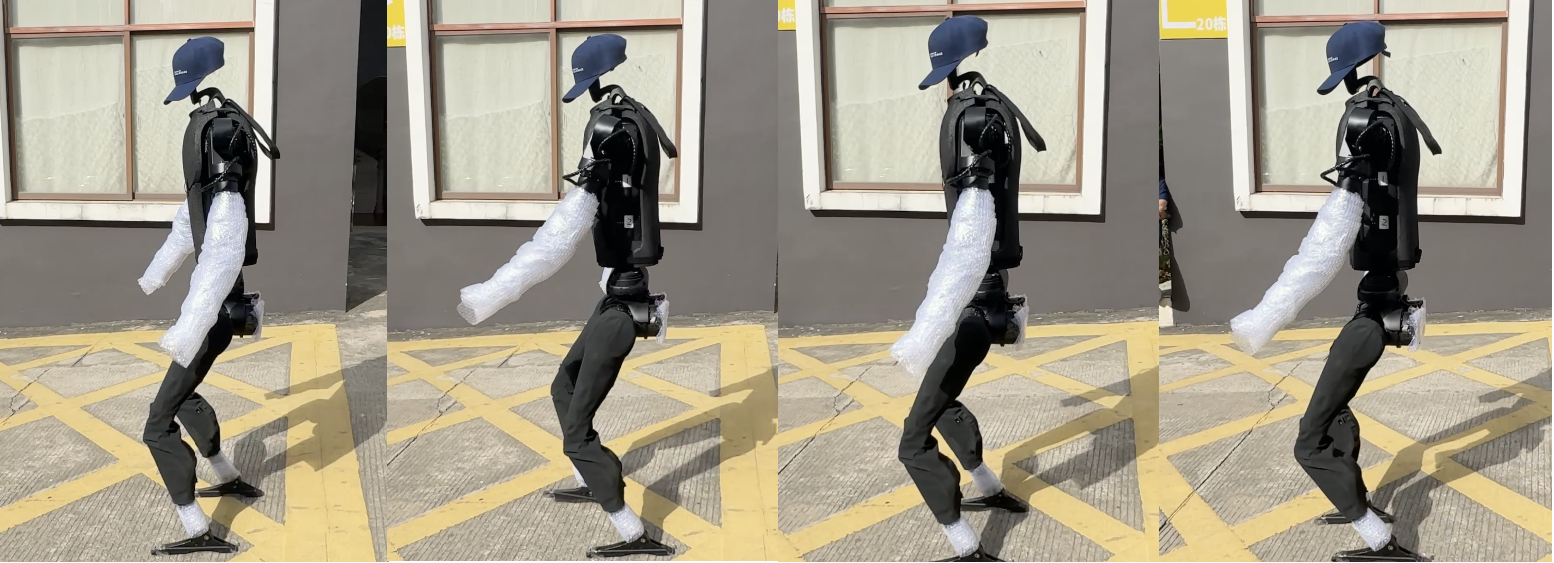}
        \caption*{(d) Descending a slope.}
    \end{minipage}
    \hfill

    \caption{Outdoor Deployment across Diverse Terrains: (a) flat ground, (b) grassy surface, (c) upward slope, and (d) downward slope.}
    \label{fig:outdoor_results}
\end{figure*}

\textbf{Effectiveness Tests in the Real World.} Figure~\ref{fig:Climb_stairs} demonstrates the robot's ability to traverse a multi-terrain course consisting of 15 cm steps, 20° slopes, and wooden boards at a speed of 0.3 m/s. Figure~\ref{fig:outdoor_results} shows the ability to stability in real-world outdoor locomotion, including flat ground, grassy surfaces, and slopes.


\begin{figure*}[htbp]
    \centering\includegraphics[width=1.0\linewidth]{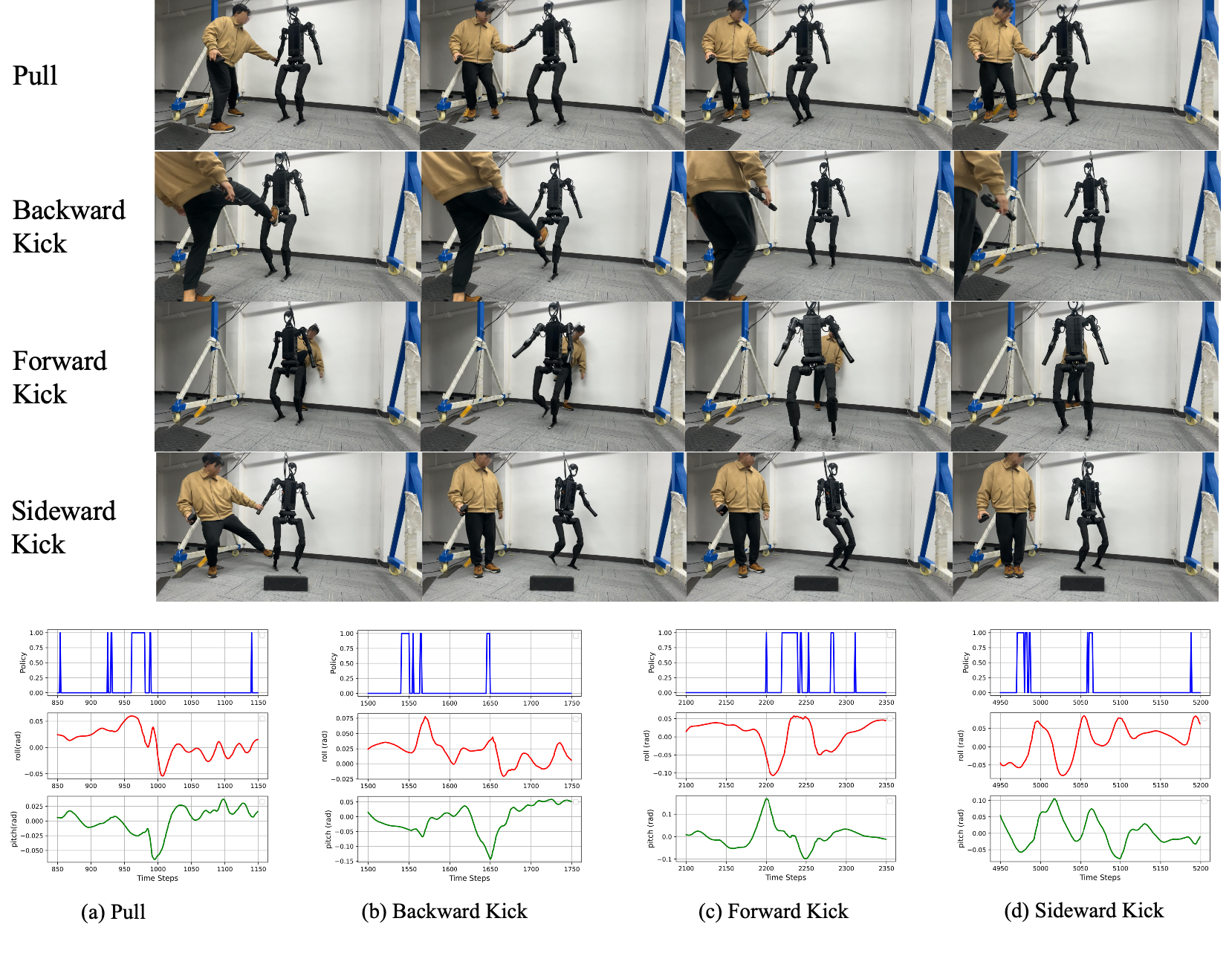}
    \caption{Disturbances in Realistic Deployment: Policy Switching value of 0 corresponds to the goal-tracking policy, while a value of 1 denotes the safety recovery policy.}
    \label{fig:Disturbance}
\end{figure*}

\begin{figure*}[htbp]
    \centering
    \includegraphics[width=0.95\linewidth]{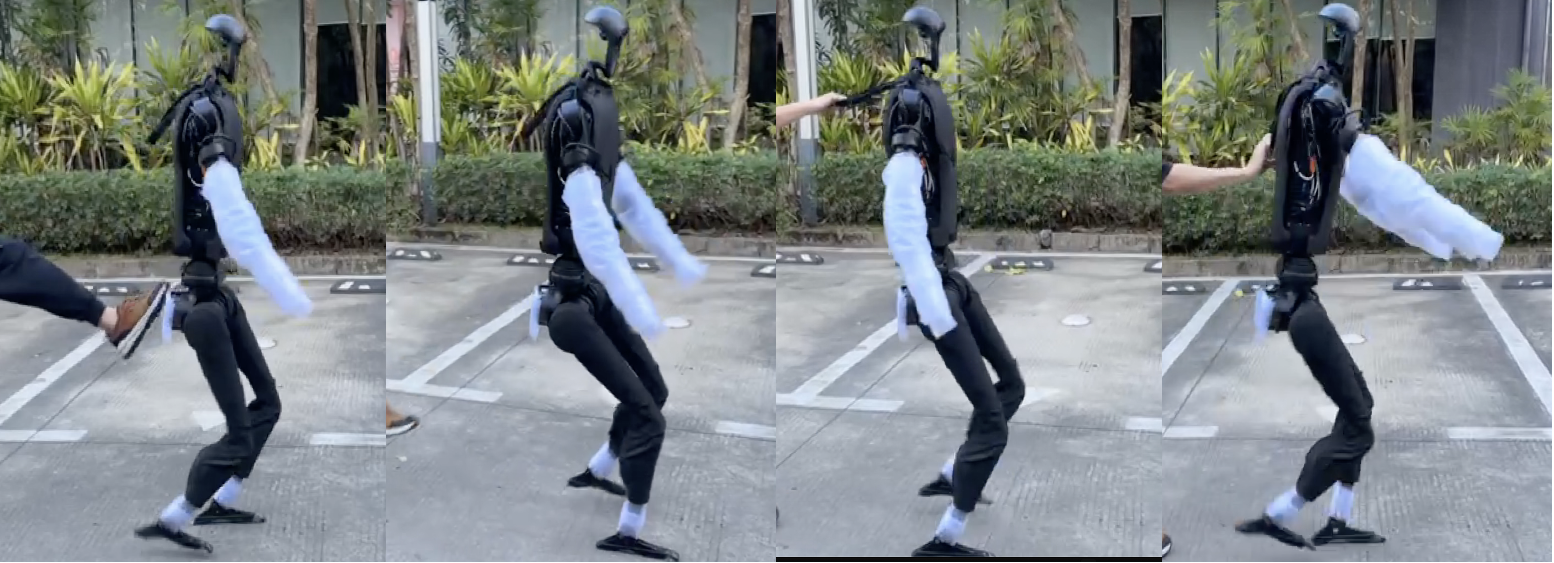}
    \caption{Robustness in Outdoor Settings: The robot responds to external disturbances in an outdoor environment by waving its arms and adjusting its gaits to maintain balance.}
    \label{fig:Outdoor_disturbance}
\end{figure*}

\textbf{Robustness Tests in the Real World.} As shown in Figure~\ref{fig:Disturbance}, we apply various disturbances to Unitree H1 lasting for one minute. Four representative disturbance segments were selected for analysis. The time is measured in control cycles, with each cycle lasting 0.01 seconds. Furthermore, we conduct an outdoor disturbance test by applying pushes, pulls, and kicks to the robot, as shown in Figure~\ref{fig:Outdoor_disturbance}. The robot waves its arms and adjusts its gaits to maintain balance. Notably, the high control frequency (100 Hz) ensures smooth transitions between policies, and isolated activations of the recovery policy have minimal impact on overall stability. 

\textbf{Case Study on Motion Tracking.} We visualize the robot's performance in tracking a punching motion. During the execution, the robot experiences an impulse impact. In response, it dynamically activates the recovery policy, switching to more conservative and safer actions. Specifically, it lowers its body, adjusts its gait, and stabilizes itself to satisfy ZMP constraints, rather than strictly following the original target motion. As a result, it generates a backtracking motion to maintain balance and ensure robustness.

\begin{figure*}[htbp]
    \centering
    \includegraphics[width=0.95\linewidth]{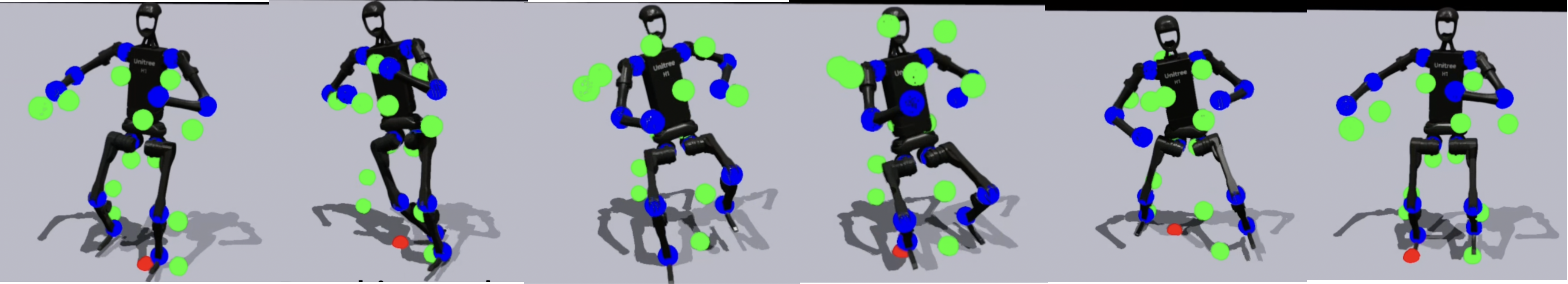}
    \caption{Robustness in Motion Tracking: The green dots represent the target keypoints, the blue dots indicate the keypoints corresponding to the DoFs, and the red dot denotes the ZMP.}
    \label{fig:Motion_recovery}
\end{figure*}

\subsection{G1 Experiments}
\label{Appendix:G1}

\begin{table}[htbp]
\centering
\caption{Comparison of Humanoid Configurations}
\begin{tabular}{lcc}
\toprule
Parameter & H1 & G1 \\
\midrule
Height & 178\,cm & 127\,cm \\
Mass   & 47\,kg  & 35\,kg \\
DoFs   & 19      & 23 \\
\bottomrule
\end{tabular}
\label{tab:humanoid_comparison}
\end{table}

G1 outperforms H1 across all three metrics (success rate, goal-tracking accuracy, and human-likeness) primarily due to the following factors:

\begin{itemize}
\item \textbf{Size and Mass:} As shown in Table~\ref{tab:humanoid_comparison}, G1’s smaller size and lighter mass contribute to lower inertial loads and reduced torque demands. This facilitates greater stability, especially during dynamic motions, and enables faster corrective actions in response to disturbances.
\item \textbf{Degrees of Freedom (DoFs):} With 23 DoFs compared to H1’s 19, G1 offers enhanced joint-level flexibility. The additional DoFs, particularly in the upper body and feet, allow for better whole-body coordination. This increased redundancy leads to improved disturbance rejection and more human-like behavior during complex locomotion tasks.
\end{itemize}

\begin{figure*}[htbp]
    \centering
    \includegraphics[width=0.95\linewidth]{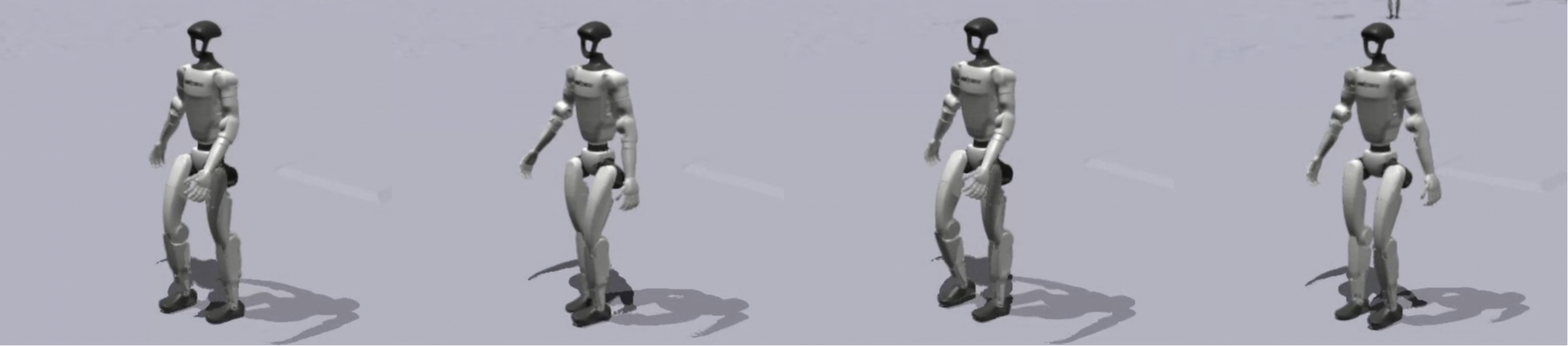}
    \caption{G1 walk with straight knees}
    \label{fig:G1}
\end{figure*}

Notably, as illustrated in Figure~\ref{fig:G1}, G1 is capable of achieving straight-knee walking, which remains a significant challenge for H1 due to its larger size and reduced agility. The additional DoFs in G1 enable finer control of limb trajectories, which is critical for maintaining balance during such biomechanically demanding gaits.

\end{document}